\documentclass[11pt,dvipsnames]{article}
\pdfoutput=1

\usepackage[activate={true,nocompatibility},final,tracking=true,kerning=true,factor=1100,stretch=10,shrink=10]{microtype}
\usepackage{graphicx}
\usepackage{booktabs}
\usepackage[numbers]{natbib}
\usepackage{amssymb}
\usepackage{amsmath}
\usepackage{subcaption}
\usepackage{multicol}
\usepackage{multirow}
\usepackage{amsthm}
\usepackage{thmtools}
\usepackage{mathtools}
\usepackage[unicode]{hyperref}
\hypersetup{
  pdfinfo={
    Title={Gradient Boosted Normalizing Flows},
    Author={Robert Giaquinto and Arindam Banerjee},
    Subject={Gradient Boosted Normalizing Flows},
    Keywords={Machine Learning, Deep Generative Model, Normalizing Flows, Gradient Boosting}
  }
}
\usepackage[ruled,vlined]{algorithm2e}
\usepackage[dvipsnames]{xcolor}
\usepackage{authblk}

\usepackage[suppress]{color-edits}
\addauthor{ab}{magenta}
\addauthor{rg}{teal}

\newcommand{\beq}{\begin{equation}}
\newcommand{\eeq}{\end{equation}}

\newcommand{\be}{\begin{eqnarray}}
\newcommand{\ee}{\end{eqnarray}}
\newcommand{\bea}{\begin{eqnarray}}
\newcommand{\eea}{\end{eqnarray}}
\newcommand{\beaa}{\begin{eqnarray*}}
\newcommand{\eeaa}{\end{eqnarray*}}
\newcommand{\bnad}{\begin{nad}}
\newcommand{\enad}{\end{nad}}

\newcommand{\x}{\mathbf{x}}

\newcommand{\z}{\mathbf{z}}

\newcommand{\cG}{{\cal G}}

\newcommand{\EE}{\mathop{\mathbb{E}}}  
\newcommand{\E}{\mathbb{E}}  

\DeclareMathOperator*{\argmin}{arg\,min}
\DeclareMathOperator*{\argmax}{arg\,max}

\newcommand{\phib}{\pmb{\phi}}

\newcommand{\vertiii}[1]{{\left\vert\kern-0.25ex\left\vert\kern-0.25ex\left\vert #1
    \right\vert\kern-0.25ex\right\vert\kern-0.25ex\right\vert}}

\def\ints{{{\rm Z} \kern -.35em {\rm Z} }}  
\def\smallints{{{\rm Z} \kern -.3em {\rm Z} }}  
\def\pints{{{\rm I} \kern -.15em {\rm N} }}      
\newcommand{\reals}{\mathbb R}

\def\cplx{{{\rm I} \kern -.45em {\rm C} }}       
\def\l2{\rm {\mathcal L}^{2}(\reals)}            

\renewcommand{\widetilde}{\tilde}

\title{Gradient Boosted Normalizing Flows}
\author{Robert Giaquinto}
\author{Arindam Banerjee}
\affil{Department of Computer Science \& Engineering\\ University of Minnesota, Twin Cities\\ Minneapolis, MN 55455, USA}
\date{}

\begin{document}
\maketitle

\begin{abstract}
By chaining a sequence of differentiable invertible transformations, normalizing flows (NF) provide an expressive method of posterior approximation, exact density evaluation, and sampling. The trend in normalizing flow literature has been to devise deeper, more complex transformations to achieve greater flexibility. We propose an alternative: Gradient Boosted Normalizing Flows (GBNF) model a density by successively adding new NF components with gradient boosting. Under the boosting framework, each new NF component optimizes a sample weighted likelihood objective, resulting in new components that are fit to the residuals of the previously trained components. The GBNF formulation results in a mixture model structure, whose flexibility increases as more components are added. Moreover, GBNFs offer a wider, as opposed to strictly deeper, approach that improves existing NFs at the cost of additional training---not more complex transformations. We demonstrate the effectiveness of this technique for density estimation and, by coupling GBNF with a variational autoencoder, generative modeling of images. Our results show that GBNFs outperform their non-boosted analog, and, in some cases, produce better results with smaller, simpler flows.%
\let\thefootnote\relax\footnote{Appearing in the 34th Conference on Neural Information Processing Systems (NeurIPS 2020), Vancouver, Canada.}
\end{abstract}

\section{Introduction}
\label{introduction}

Deep generative models seek rich latent representations of data, and provide a mechanism for sampling new data. Beyond their wide-ranging applications, generative models are an attractive class of models that place strong assumptions on the data and hence exhibit higher asymptotic bias when the model is incorrect \citep{banerjee_analysis_2007}. A popular approach to generative modeling is with variational autoencoders (VAEs) \citep{kingma_autoencoding_2014}. A major challenge in VAEs, however, is that they assume a factorial posterior, which is widely known to limit their flexibility \citep{rezende_variational_2015,kingma_improving_2016,miller_variational_2017,chen_variational_2017,huang_improving_2018,tomczak_vae_2018,casale_gaussian_2018,vandenberg_sylvester_2018}. Further, VAEs do not offer exact density estimation, which is a requirement in many settings.

Normalizing flows (NF) are an important recent development and can be used in both density estimation \citep{tabak_family_2013,rippel_highdimensional_2013,dinh_nice_2015} and variational inference \citep{rezende_variational_2015}. Normalizing flows are smooth, invertible transformations with tractable Jacobians, which can map a complex data distribution to simple distribution, such as a standard normal \citep{papamakarios_normalizing_2019}. In the context of variational inference, a normalizing flow transforms a simple, known base distribution into a more faithful representation of the true posterior. As such, NFs complement VAEs, providing a method to overcome the limitations of a factorial posterior. Flow-based models are also an attractive approach for density estimation \citep{tabak_family_2013,dinh_nice_2015,dinh_density_2017,papamakarios_masked_2017,decao_block_2019,grathwohl_ffjord_2019,salman_deep_2018,ho_flow_2019,huang_neural_2018,kingma_glow_2018,papamakarios_normalizing_2019} because they provide exact density computation and sampling with only a single neural network pass (in some instances) \citep{durkan_neural_2019}. 

Recent developments in NFs have focused of creating deeper, more complex transformations in order to increase the flexibility of the learned distribution \citep{kingma_glow_2018,ma_macow_2019,ho_flow_2019,chen_vflow_2020,huang_neural_2018,chen_residual_2019,behrmann_invertible_2019}. With greater model complexity comes a greater risk of overfitting while slowing down training, prediction, and sampling. Boosting \citep{freund_decisiontheoretic_1997,mason_boosting_1999,friedman_additive_2000,friedman_greedy_2001,friedman_stochastic_2002} is flexible, robust to overfitting, and generally one the most effective learning algorithms in machine learning \citep{hastie_elements_2001}. While boosting is typically associated with regression and classification, it is also applicable in the unsupervised setting \citep{rosset_boosting_2002,campbell_universal_2019,miller_variational_2017,guo_boosting_2016,grover_boosted_2018,campbell_universal_2019,locatello_boosting_2018}.

\paragraph{Our contributions.} In this work we propose a \emph{wider}, as opposed to strictly deeper, approach for increasing the expressiveness of density estimators and posterior approximations. Our approach, \emph{gradient boosted normalizing flows} (GBNF), iteratively adds new NF components to a model based on gradient boosting, where each new NF component is fit to the residuals of the previously trained components. A weight is learned for each component of the GBNF model, resulting in a mixture structure. However, unlike a mixture model, GBNF offers the optimality advantages associated with boosting \citep{bartlett_convexity_2006}, and a simplified training objective that focuses solely on optimizing a single new component at each step. GBNF compliments existing flow-based models, improving performance at the cost of additional training cycles---not more complex transformations. Prediction and sampling are not slowed with GBNF, as each component is independent and operates in parallel.

While gradient boosting is straight-forward to apply in the density estimation setting, our analysis highlights the need for \emph{analytically} invertible flows in order to efficiently boost flow-based models for variational inference. Further, we address the ``decoder shock'' phenomenon---a challenge unique to VAEs with GBNF approximate posteriors, where the loss increases suddenly coinciding with the introduction of a new component. Our results show that GBNF improves performance on density estimation tasks, capable of modeling multi-modal data. Lastly, we augment the VAE with a GBNF variational posterior, and present image modeling results on par with state-of-the-art NFs. 

The remainder of the paper is organized as follows. In Section \ref{sec:background} we briefly review normalizing flows. In Section \ref{sec:gbnf_exact} we introduce GBNF for density estimation, and Section \ref{sec:gbnf_approximate} we extend our idea for the approximate inference setting. In Section \ref{sec:discussion} we discuss normalizing flows that are compatible with GBNF, and the ``decoder shock'' phenomenon. In Section \ref{sec:experiments} we present results. Finally, we conclude the paper in Section \ref{sec:conclusion}.

\section{Background}
\label{sec:background}

Normalizing flows are applicable to both approximate density estimation techniques, like variational inference, as well as tractable density estimation. As such, we review variational inference techniques using modern deep neural networks to amortize the cost of learning parameters. We then share how augmenting deep generative models with normalizing flows improves variational inference. Finally, we highlight recent work using normalizing flows as density estimators directly on the data.

\subsection{Variational Inference}
Approximate inference plays an important role in fitting complex probabilistic models. Variational Inference (VI), in particular, transforms inference into an optimization problem with the goal of finding a variational distribution $q_\phi(\z \mid \x)$ that closely approximates the true posterior $p(\z \mid \x)$, where $\x$ are the observed data, $\z$ the latent variables, and $\phi$ are learned parameters \citep{jordan_introduction_1999,wainwright_graphical_2007,blei_variational_2017}. Writing the log-likelihood of the data in terms of the approximate posterior reveals:
\begin{align}
\label{eq:lhood}
&\log p_\theta(\x) = \underbrace{\E_{q_\phi} \left[ \log \left[ \frac{p_\theta(\x, \z)}{q_\phi(\z \mid \x)} \right] \right]}_{\mathcal{L}_{\theta, \phi}(\x) \quad \text{(ELBO)}} + \underbrace{\E_{q_\phi} \left[ \log \left[ \frac{q_\phi(\z \mid \x)}{p_\theta(\z \mid \x)} \right] \right]}_{KL(q_\phi(\z \mid \x) \, || \, p_\theta(\z \mid \x))}
\end{align}
Since the second term in \eqref{eq:lhood} is the Kullback-Leibler (KL) divergence, which is non-negative, then the first term forms a lower bound on the log-likelihood of the data, and hence referred to as the evidence lower bound (ELBO).

\subsection{Variational Autoencoder}
\citet{kingma_autoencoding_2014,rezende_stochastic_2014} show that a re-parameterization of the ELBO can result in a differentiable bound that is amenable to optimization via stochastic gradients and back-propagation. Further, \citet{kingma_autoencoding_2014} structure the inference problem as an autoencoder, introducing the variational autoencoder (VAE) and minimizing the negative-ELBO $\mathcal{F}_{\phi, \theta}^{(VI)}(\x)$. Re-writing the VI objective $\mathcal{F}_{\phi, \theta}^{(VI)}(\x)$ as:
\begin{equation}
\label{eq:negative_elbo}
\mathcal{F}_{\phi, \theta}^{(VI)}(\x) = \E_{q_\phi} \left[ -\log p_\theta(\x \mid \z) \right] +  KL\left( q_\phi(\z \mid \x)  \, || \, p(\z)  \right)~,
\end{equation}
shows the probabilistic decoder $p_\theta(\x \mid \z)$, and highlights how the VAE encodes the latent variables $\z$ with the variational posterior $q_\phi(\z \mid \x)$, but $q_\phi(\z \mid \x)$ is regularized with the prior $p(\z)$.

\subsection{Normalizing Flows}
\citet{tabak_density_2010,tabak_family_2013} introduce normalizing flows (NF) as a composition of simple maps. Parameterizing flows with deep neural networks \citep{dinh_density_2017,dinh_nice_2015,rippel_highdimensional_2013} has popularized the technique for density estimation and variational inference \cite{papamakarios_normalizing_2019}.

\paragraph{Variational Inference} \citet{rezende_variational_2015} use NFs to modify the VAE's \citep{kingma_autoencoding_2014} posterior approximation $q_0$ by applying a chain of $K$ transformations $\z_K = f_K \circ \dots \circ f_1(\z_0)$ to the inference network output $\z_0 \sim q_0(\z_0 \mid \x)$. By defining $f_k, k=1,\ldots,K$ as an invertible, smooth mapping, by the chain rule and inverse function theorem $z_k = f_k(z_{k-1})$ has a computable density \citep{tabak_density_2010,tabak_family_2013}:
\begin{align}
\label{eq:nf_density}
q_k(\z_k) = q_{k-1}(\z_{k-1}) \left| \det \frac{\partial f_k^{-1}}{\partial \z_{k-1}} \right| = q_{k-1}(\z_{k-1}) \left| \det \frac{\partial f_k}{\partial \z_{k-1}} \right|^{-1} ~.
\end{align}
Thus, a VAE with a $K$-step flow-based posterior minimizes the negative-ELBO:
\begin{align}
\label{eq:nf_elbo}
\mathcal{F}_{\theta, \phi}^{(VI)}(\x) &= \E_{q_0} \left[ -\log p_{\theta}(\x \mid \z_K) - \sum_{k=1}^K \log \left| \det \frac{\partial f_k}{\partial \z_{k-1}} \right| \right] + KL \left( q_0(\z_0 \mid \x) \, || \, p(\z_K) \right)~, 
\end{align}
where $q_0(\z_0 \mid \x)$ is a known base distribution (e.g. standard normal) with parameters $\phi$.

\paragraph{Density Estimation}
Given a set of samples $\{ \x_i \}_{i=1}^n$ from a target distribution $p^*$, our goal is to learn a flow-based model $p_\phi(\x)$, which corresponds to minimizing the forward KL-divergence: $\mathcal{F}^{(ML)}(\phi) = KL(p^*(\x) \, || \, p_\phi(\x))$ \citep{papamakarios_normalizing_2019}. A NF formulates $p_\phi(\x)$ as a transformation $\x = f(\z)$ of a base density $p_0(\z)$ with $f = f_K \circ \dots \circ f_1$ as a $K$-step flow \citep{dinh_nice_2015,dinh_density_2017,papamakarios_masked_2017}. Thus, to estimate the expectation over $p^*$ we take a Monte Carlo approximation of the forward KL, yielding:
\begin{align}
\label{eq:density_estimation}
\mathcal{F}^{(ML)}(\phi) &\approx - \frac{1}{n} \sum_{i=1}^n \left[ \log p_0 \left( f^{-1}(\x_i) \right) + \sum_{k=1}^K \log \left| \det \frac{\partial f_k^{-1}}{\partial \x_i}   \right|  \right] ~,
\end{align}
which is equivalent to fitting the model to samples $\{ \x_i \}_{i=1}^n$ by maximum likelihood estimation \citep{papamakarios_normalizing_2019}.

\subsection{Gradient Boosting}
Gradient boosting \citep{mason_boosting_1999,friedman_additive_2000,friedman_greedy_2001,friedman_stochastic_2002} considers the minimization of a loss $\mathcal{F}(G)$, where $G(\cdot)$ is a function representing the current model. Consider an additive perturbation around $G$ to $G + \epsilon g$, where $g$ is a function representing a new component. A Taylor expansion as $\epsilon \rightarrow 0$:
\begin{equation}
\label{eq:boosting}
\mathcal{F}(G + \epsilon g) = \mathcal{F}(G) + \epsilon \langle g, \nabla \mathcal{F}(G) \rangle + o(\epsilon^2)~,
\end{equation}
reveals the functional gradient $\nabla \mathcal{F}(G)$, which is the direction that reduces the loss at the current solution.

Thus, to minimize loss $\mathcal{F}(G)$ at the current model, choose the best function $g$ in a class of functions $\mathcal{G}$ (e.g. regression trees), which corresponds to solving a linear program where $\nabla \mathcal{F}(G)$ defines the weights for every function in $\mathcal{G}$. Underlying Gradient boosting is a connection to conditional gradient descent and the Frank-Wolfe algorithm \citep{frank_algorithm_1956}: we first solve a constrained convex minimization problem to choose $g$, then solve a line-search problem to appropriately weight $g$ relative to the previous components $G$ \citep{campbell_universal_2019,guo_boosting_2016}.

\section{Density Estimation with GBNF}
\label{sec:gbnf_exact}

\newcommand{\widefigwidth}{0.24\linewidth}
\begin{figure}[t]
\centering

\begin{subfigure}[b]{\widefigwidth}
\centering
\caption*{\textbf{Target}}
\vspace{-2mm}
\includegraphics[width=\linewidth]{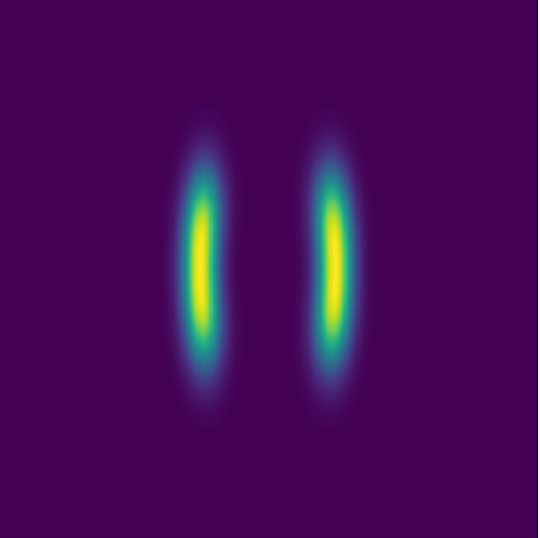}
\end{subfigure}%
\hspace{1mm}%
\begin{subfigure}[b]{\widefigwidth}
\centering
\caption*{\centering \textbf{1 Component}}
\vspace{-2mm}
\includegraphics[width=\linewidth]{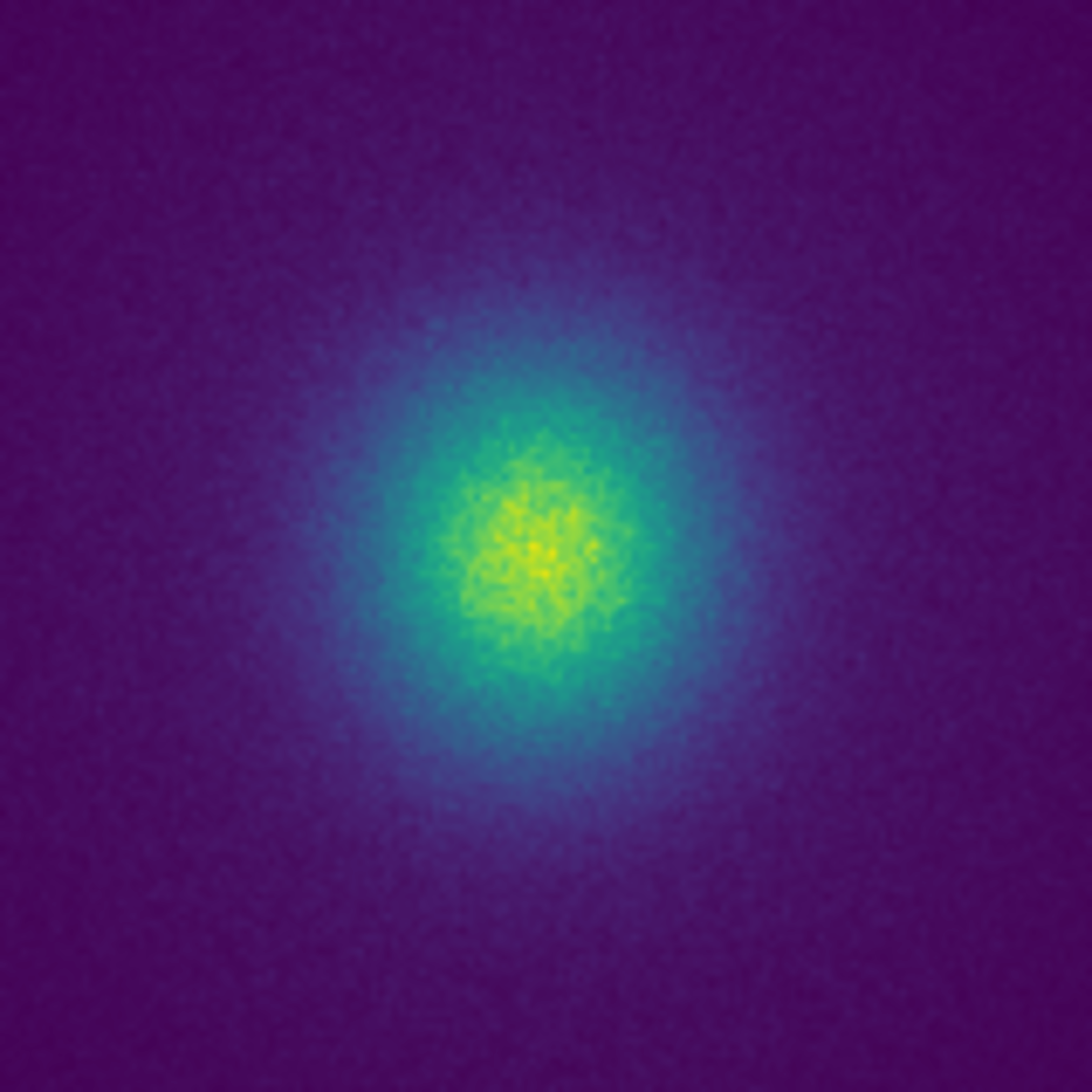}
\end{subfigure}%
\hspace{1mm}%
\begin{subfigure}[b]{\widefigwidth}
\centering
\caption*{\centering \textbf{2 Components}}
\vspace{-2mm}
\includegraphics[width=\linewidth]{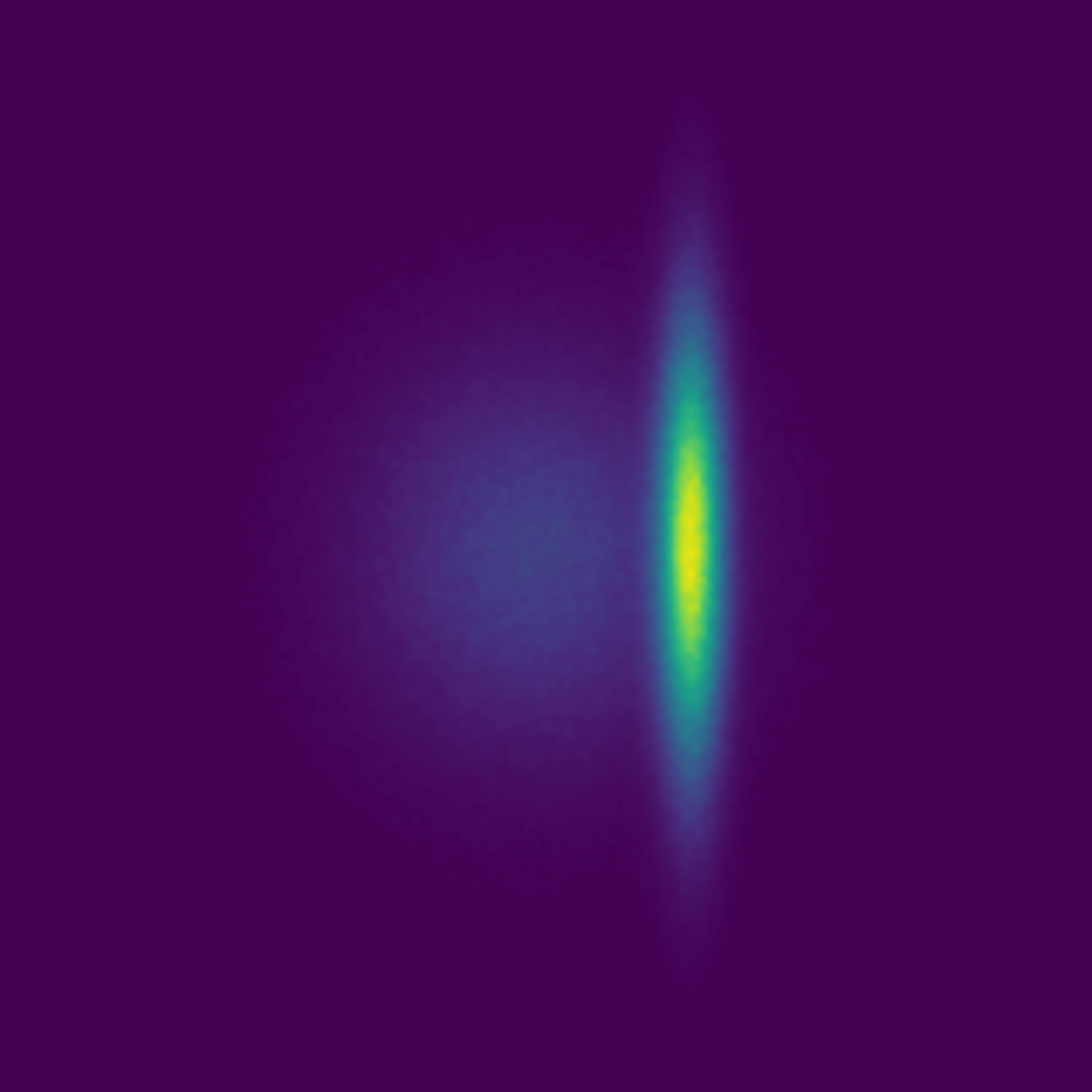}
\end{subfigure}%
\hspace{1mm}%
\begin{subfigure}[b]{\widefigwidth}
\centering
\caption*{\centering \textbf{Fine-Tune}}
\vspace{-2mm}
\includegraphics[width=\linewidth]{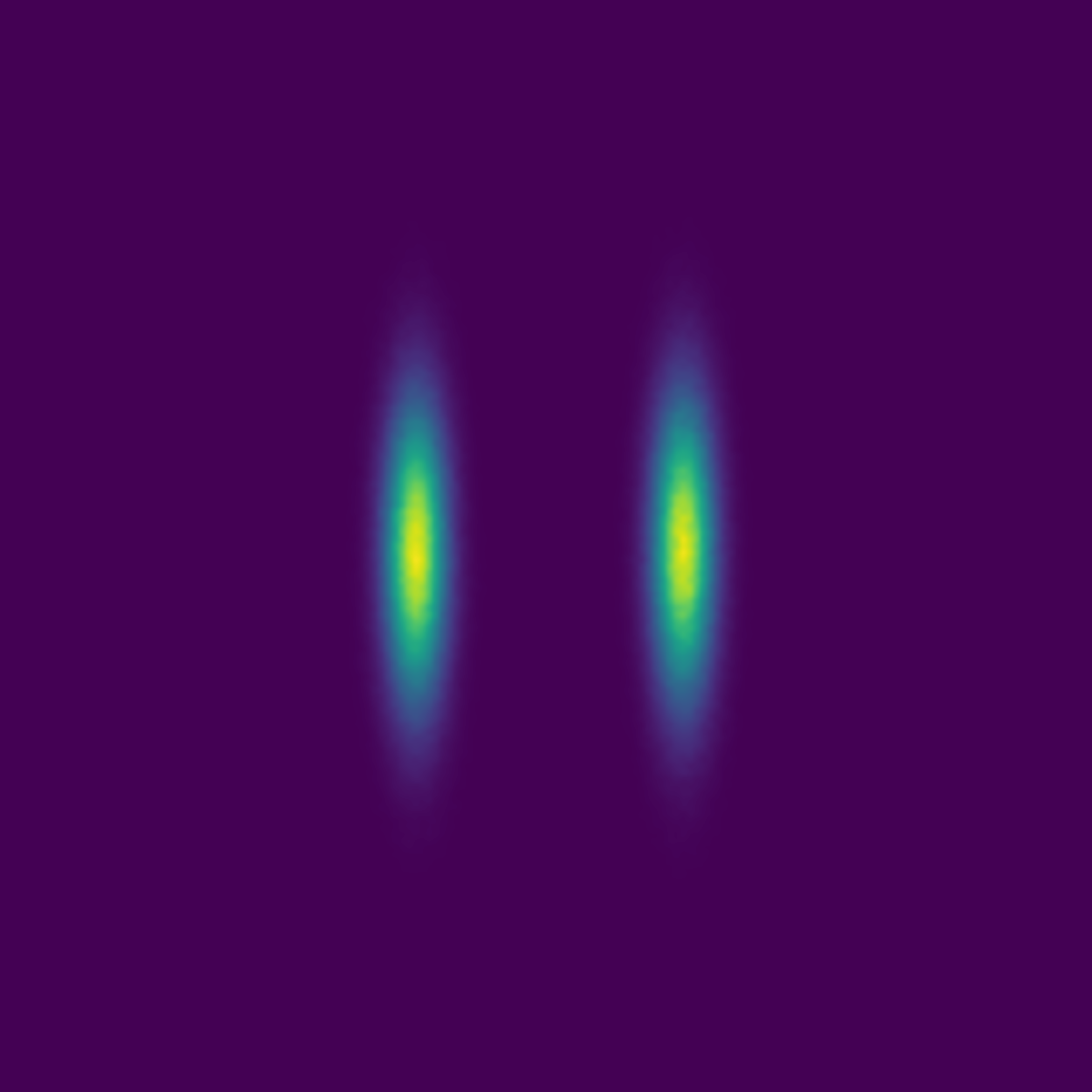}
\end{subfigure}%

\caption{Example of GBNF: A simple affine flow (one scale and shift operation) cannot model the data distribution and leads to mode-covering (\textbf{1 Component}).
In \textbf{2 Components}, GBNF introduces a second component, which seeks a region of high probability that is not well modeled by the first component. Here, fine-tuning components with additional \emph{boosted} training leads to a better solution, shifting the first component to the left ellipsoid and re-weighing appropriately as shown in \textbf{Fine-Tune}.}
\label{fig:gbnf_toy}
\end{figure}

\emph{Gradient boosted normalizing flows} (GBNF) build on recent ideas in boosting for variational inference \citep{guo_boosting_2016,miller_variational_2017} and generative models \citep{grover_boosted_2018} in order to increase the flexibility of density estimators and posteriors approximated with NFs. A GBNF is constructed by successively adding new components based on forward-stagewise gradient boosting, where each new component $g^{(c)}_{K}$ is a $K$-step normalizing flow that is fit to the functional gradient of the loss from the $(c-1)$ previously trained components $G^{(c-1)}_{K}$.

Gradient boosting assigns a weight $\rho_c$ to the new component $g^{(c)}_{K}$ and we restrict $\rho_c \in [0,1]$ to ensure the model stays a valid probability distribution. The resulting density can be written as a mixture model:
\begin{align}
G_{K}^{(c)}(\x) &= \psi\left((1-\rho_c) \psi^{-1}(G_{K}^{(c-1)}(\x)) + \rho_c \psi^{-1}( g_{K}^{(c)}(\x))\right)/\Gamma_{(c)} ~,
\label{eq:gbnf_de_model}
\end{align}
where the full model $G_{K}^{(c)}(\x)$ is a monotonic function $\psi$ of a convex combination of fixed components  $G_{K}^{(c-1)}$ and new component $g_{K}^{(c)}$, and $\Gamma_{(c)}$ is the partition function. Two special cases are of interest: first, when $\psi(a) = a$, which corresponds to a standard additive mixture model with $\Gamma_{(c)} = 1$:
\begin{align*}
\log G_{K}^{(c)}(\x) &= \log \left((1-\rho_c) (G_{K}^{(c-1)}(\x)) + \rho_c ( g_{K}^{(c)}(\x))\right) ~.
\end{align*}
Second, when $\psi(a) = \exp(a)$ with $\psi^{-1}(a) = \log(a)$, which corresponds to a multiplicative mixture model \citep{grover_boosted_2018,cranko_boosting_2019}:
\begin{align*}
\log G_{K}^{(c)}(\x) &= \left((1-\rho_c) \log (G_{K}^{(c-1)}(\x)) + \rho_c \log ( g_{K}^{(c)}(\x))\right) - \log \Gamma_{(c)} ~.
\end{align*}
The advantage of GBNF over a standard mixture model with a pre-determined fixed number of components is that additional components can always be added to the model, and the weights $\rho_c$ for non-informative components will degrade to zero. Since each GBNF component is a NF, we evaluate \eqref{eq:gbnf_de_model} recursively, and at each stage the last component is computed by the change of variables formula:
\begin{equation}
g_K^{(c)} = p_{0} \left( f_{c}^{-1}(\x) \right) \prod_{k=1}^K  \left| \det \frac{ \partial f^{-1}_{k,c} }{ \partial \x } \right| ~,
\end{equation}
where $f_c = f_{c,K} \circ \cdots \circ f_{c,1}$ is the $K$-step flow transformation for component $c$, and the base density $p_0$ is shared by all components. In our formulation we consider $c$ components, where $c$ is fixed and finite.

\paragraph{Density Estimation}
With GBNF density estimation is similar to \eqref{eq:density_estimation}: we seek flow parameters $\phib = [\phi_1, \dots, \phi_c]$ that minimize $KL\left( p^*(\x) \, || \, G_{K}^{(c)}(\x) \right)$, which for finite number of samples $\{ \x_i\}$ drawn from $p^*(\x)$ corresponds to minimizing:
\begin{equation}
    \label{eq:exact_gbnf_objective}
    \mathcal{F}^{(ML)}(\phib) = - \frac{1}{n} \sum_{i=1}^n \left[ \log \left\{ \psi \left( (1 - \rho_c) \psi^{-1}(G_{K}^{(c-1)}(\x_i)) + \rho_c \psi^{-1}(g_{K}^{(c)}(\x_i))  \right)/\Gamma_{(c)}\right\} \right]~.
\end{equation}
Directly optimizing \eqref{eq:exact_gbnf_objective} for mixture model $G_{K}^{(c)}$ is non-trivial. Gradient boosting, however, provides an elegant solution that greatly simplifies the problem. During training, the first component is fit using a traditional objective function---no boosting is applied\footnote{No boosting during the first stage is equivalent to setting $G^{(0)}_{K}(\x)$ to uniform on the domain of $\x$.}. At stages $c>1$, we already have $G_{K}^{(c-1)}$, consisting of a convex combination of the $(c-1)$ $K$-step flow models from the previous stages, and we train a new component $g_{K}^{(c)}$ by taking a Frank-Wolfe \citep{frank_algorithm_1956,beygelzimer_online_2015,campbell_universal_2019,jaggi_revisiting_2013,locatello_boosting_2018,guo_boosting_2016} linear approximation of the loss \eqref{eq:exact_gbnf_objective}. Since jointly optimizing w.r.t. both $g_{K}^{(c)}$ and $\rho_c$ is a challenging non-convex problem \citep{guo_boosting_2016}, we train $g_{K}^{(c)}$  until convergence, and then use \eqref{eq:exact_gbnf_objective} as the objective to optimize w.r.t the corresponding weight $\rho_c$.

\subsection{Updates to New Boosting Components}
\label{sec:gbnf_de_update}

\paragraph{Additive Boosting}
We first consider the special case $\psi(a) = a$ and $\Gamma_{(c)}=1$, corresponding to the additive mixture model. Our goal is to derive an update to the new component $g_{K}^{(c)}$ using functional gradient descent. Thus, we take the gradient of \eqref{eq:exact_gbnf_objective} w.r.t. fixed parameters $\phib_{1:c-1}$ of $G_{K}^{(c)}$ at $\rho_c \rightarrow 0$, giving:
\small
\begin{align}
\label{eq:exact_gradient}
\nabla_{\phib_{1:c-1}} \mathcal{F}^{(ML)}(\phib) \Big|_{\rho_c \rightarrow 0} &= - \frac{ 1 - \rho_c }{ (1 - \rho_c) G_{K}^{(c-1)} + \rho_c g_{K}^{(c)} }  \Bigg|_{\rho_c \rightarrow 0} = - \frac{ 1 }{ G_{K}^{(c-1)} } 
\end{align}
\normalsize
Since $G_{K}^{(c-1)}$ is fixed, then maximizing $-\mathcal{F}^{(ML)}(\phib)$ is achieved by choosing a new component $g_{K}^{(c)}$ and weighing by the negative of the gradient from \eqref{eq:exact_gradient} over the samples: 
\begin{align}
\label{eq:de_update}
g_{K}^{(c)} &= \argmax_{g_{K} \in \mathcal{G}_{K}}  \frac{1}{n} \sum_{i=1}^n \frac{ g_{K}(\x_i) }{ G_{K}^{(c-1)}(\x_i) } ~,
\end{align}
where $\mathcal{G}_{K}$ is the family of $K$-step flows. Note that \eqref{eq:de_update} is a linear program in which $G_K^{(c-1)}(\x_i)$ is a constant, and will hence yield a degenerate point probability distribution where the entire probability mass is placed at the minimum point of $G_K^{(c-1)}$. To avoided the degenerate solution, a standard approach adds an entropy regularization term which is controlled by the hyperparameter $\lambda$ \citep{guo_boosting_2016,campbell_universal_2019}:
\begin{align}
\label{eq:de_additive_update_entropy}
g_{K}^{(c)} &= \argmax_{g_{K} \in \mathcal{G}_{K}}  \frac{1}{n} \sum_{i=1}^n \frac{ g_{K}(\x_i) }{ G_{K}^{(c-1)}(\x_i) } - \lambda \sum_{i=1}^n g_K(\x_i) \log g_K(\x_i) ~.
\end{align}

\paragraph{Multiplicative Boosting}
In this paper, we instead use $\psi(a) = \exp(a)$ with $\psi^{-1}(a)=\log(a)$, which corresponds to the multiplicative mixture model, and, from the boosting perspective, a multiplicative boosting model \cite{grover_boosted_2018,cranko_boosting_2019}. However, in contrast to the existing literature on multiplicative boosting for probabilistic models, we consider boosting with normalizing flow components. In the multiplicative setting, explicitly maintaining the convex combination between $G_K^{(c-1)}$ and $g_K^{(c)}$ is unnecessary: the partition function $\Gamma_{(c)}$ ensures the validity of the probabilistic model. Thus, the multiplicative GBNF seeks a new component $g_K^{(c)}$ that minimizes:
\begin{equation}
   \mathcal{F}^{(ML)}(\phib) = - \frac{1}{n} \sum_{i=1}^n \left[  \left(  \log(G_{K}^{(c-1)}(\x_i)) + \rho_c \log(g_{K}^{(c)}(\x_i))  \right) - \log \Gamma_{(c)} \right]~.
\label{eq:multi}
\end{equation}
The partition function is defined as $\Gamma_{(c)} = \int_{x} \prod_{j=1}^c (g_K^{(j)})^{\rho_j}(\x) p_0(\x) d\x$ and computing $\Gamma_{(c)}$ for GBNF is straightforward since normalizing flows learn self-normalized distributions---and hence can be computed without resorting to simulated annealing or Markov chains \citep{grover_boosted_2018}. Moreover, following standard properties \citep{grover_boosted_2018,cranko_boosting_2019}, we also inherit a recursive property of the partition function. To see the recursive property, denote the un-normalized GBNF density as $\widetilde{G}_K^{(c)}$, where $\widetilde{G}_K^{(c)} \propto G_K^{(c)}$ but $\widetilde{G}_K^{(c)}$ does not integrate to $1$. Then, by definition:
\begin{equation*}
    \Gamma_{(c)} G_K^{(c)}(\x) = g_K^{(c)}(\x)^{\rho_c} \widetilde{G}_K^{(c-1)}(\x) = \Gamma_{(c-1)} g_K^{(c)}(\x)^{\rho_c} G_K^{(c-1)}(\x)~,
\end{equation*}
then, integrating both sides and using $\int_{\x} G_K^{(c)}(\x) d\x = 1$ gives
\begin{equation*}
    \Gamma_{(c)} = \Gamma_{(c-1)} \int_{\x} g_K^{(c)}(\x)^{\rho_c} G_K^{(c-1)}(\x) d\x = \Gamma_{(c-1)} \E_{G_K^{(c-1)}}\left[ g_K^{(c)}(\x)^{\rho_c} \right]~.
\end{equation*}
and therefore $\Gamma_{(c)} = \Gamma_{(c-1)} \E_{G_K^{(c-1)}}[ g_K^{(c)}(\x)^{\rho_c}]$, as desired.

The objective in \eqref{eq:multi} represents the loss under the model $G_K^{(c)}$ which followed from minimizing the forward KL-divergence $KL( p^* \| \mathbf{G}_K^{(c)} )$, where $\mathbf{G}_K^{(c)}$ is the normalized approximate distribution and $p^*$ the target distribution. To improve \eqref{eq:multi} with gradient boosting, consider the difference in losses after introducing a new component $g_K^{(c)}$ to the model: 
\begin{align}
\label{eq:kl_difference}
KL( p^* \| \mathbf{G}_K^{(c-1)} ) - KL( p^* \| \mathbf{G}_K^{(c)} )  &= \E_{p^*} \left[ \log \frac{ p^*(\x) }{ G_K^{(c-1)}(\x) } - \log \frac{ p^*(\x) }{ G_K^{(c)}(\x) } \right] \nonumber \\
& = \E_{p^*} \left[ \log \frac{ G_K^{(c)}(\x) }{ G_K^{(c-1)}(\x) } \right] \nonumber \\
& = \E_{p^*} \left[ \log \frac{ (g_K^{(c)}(\x))^{\rho_c} \widetilde G_K^{(c-1)}(\x) }{ \Gamma_{c-1} \E_{G_K^{(c-1)}}[(g_K^{(c)}(\x))^{\rho_c}]} \times \frac{ \Gamma_{c-1} }{ \widetilde G_K^{(c-1)}(\x) } \right] \nonumber \\
& = \E_{p^*} \left[ \log \frac{ (g_K^{(c)}(\x))^{\rho_c} }{ \E_{G_K^{(c-1)} }[(g_K^{(c)}(\x))^{\rho_c}]}\right]  \nonumber \\
& = \E_{p^*} \left[ \log g_K^{(c)}(\x))^{\rho_c}  \right] - \log \E_{G_K^{(c-1)}} \left[ g_K^{(c)}(\x))^{\rho_c} \right] \nonumber \\
& \stackrel{(a)}{\geq} \E_{p^*} \left[ \log g_K^{(c)}(\x))^{\rho_c}  \right] - \log \left(\E_{G_K^{(c-1)}}[g_K^{(c)}(\x)] \right)^{\rho_c} \nonumber \\
& = \rho_c \left\{ \E_{p^*} \left[ \log g_K^{(c)}(\x)  \right] - \log \E_{G_K^{(c-1)}} \left[ g_K^{(c)}(\x) \right] \right\}~,
\end{align}
where (a) follows by Jensen's inequality since $\rho_c \in [0,1]$. Note that we want to choose the new component $g_K^{(c)}(\x)$ so that $KL( p^* \| \mathbf{G}_K^{(c)})$ is minimized, or equivalently, for a fixed $\mathbf{G}_K^{(c-1)}$, the difference $KL(p^* \| \mathbf{G}_K^{(c-1)} ) - KL(p^* \| \mathbf{G}_K^{(c)})$ is maximized. Since $\rho_c \geq 0$, it suffices to focus on the following maximization problem:
\begin{equation}
\label{eq:multiplicative_update}
g_K^{(c)} = \argmax_{g_K \in \mathcal{G}_K}~ \E_{p^*} \left[ \log g_K(\x)  \right] - \log \E_{G_K^{(c-1)}} \left[ g_K(\x) \right ] ~.
\end{equation}

If we choose a new component according to:
\begin{equation*}
g_K^{(c)}(\x) = \frac{ p^*(\x) }{ G_K^{(c-1)}(\x) }~,
\end{equation*}
then, with this choice of $g_K^{(c)}$, we see that \eqref{eq:multiplicative_update} reduces to:
\begin{equation*}
\E_{p^*} \left[ \log \frac{ p^*(\x) }{ G_K^{(c-1)}(\x) } \right] - \log \E_{G_K^{(c-1)}} \left[ \frac{ p^*(\x) }{ G_K^{(c-1)}(\x) } \right ] = KL \left( p^* \| \mathbf{G}_K^{(c-1)} \right) - \underbrace{\log \E_{p^*(\x)} \left[ 1 \right]}_{0} ~.
\end{equation*}
Our choice of $g_K^{(c)}$ is, therefore, optimal and gives a lower bound to \eqref{eq:kl_difference} with $ KL( p^* \| \mathbf{G}_K^{(c)}) \rightarrow 0$. The solution to \eqref{eq:multiplicative_update} can also be understood in terms of the change-of-measure inequality \citep{banerjee_bayesian_2006,donsker_principal_1976}, which also forms the basis of the PAC-Bayes bound and a certain regret bounds \citep{banerjee_bayesian_2006}.

%
%
Similar to the additive case, the new component chosen in \eqref{eq:multiplicative_update} shows that $g_K^{(c)}$ maximizes the likelihood of the samples while discounting those that are already explained by $G_K^{(c-1)}$. Unlike the additive GBNF update in \eqref{eq:de_additive_update_entropy}, however, the multiplicative GBNF update is a numerically stable and does not require an entropy regularization term.

%
%
Further, our analysis reveals the source of a surrogate loss function \citep{nguyen_divergences_2005,bartlett_convexity_2006,nguyen_surrogate_2009} which optimizes the global objective---namely, when written as as a minimization \eqref{eq:multiplicative_update} is the \emph{weighted} negative log-likelihood of the samples. Surrogate loss functions are common in the boosting framework \citep{freund_decisiontheoretic_1997,schapire_boosting_1998,bartlett_convexity_2006,friedman_additive_2000,freund_experiments_1996,buhlmann_boosting_2007,tolstikhin_adagan_2017,rosset_boosting_2002}. Adaboost \citep{freund_experiments_1996,freund_decisiontheoretic_1997}, in particular, solves a re-weighted classification problem where weak learners, in the form of decision trees, optimize surrogate losses like information gain or Gini index. The negative log-likelihood is specifically chosen as a surrogate loss function in other boosted probabilistic and density estimation models which also have $f$-divergence based global objectives \citep{grover_boosted_2018,rosset_boosting_2002}, however here we clarify that the surrogate loss follows from \eqref{eq:kl_difference}.

%
%
In practice, instead of weighing the loss by to the reciprocal of the fixed components, we follow \citet{grover_boosted_2018} and train the new component to perform maximum likelihood estimation over a re-weighted data distribution:
\begin{align}
\label{eq:multiplicative_update_expectation}
g_K^{(c)} = \argmin_{g_K \in \mathcal{G}_K} \, \EE_{\mathcal{D}^{(c-1)}} \left[ - \log g_K \right]
\end{align}
where $\mathcal{D}^{(c-1)}$ denotes a re-weighted data distribution whose samples are drawn with replacement using sample weights inversely-proportional to $G_K^{(c-1)}$. Since \eqref{eq:multiplicative_update_expectation} is the same objective as the generative multiplicative boosting model in \citet{grover_boosted_2018}, where we have set their parameter $\beta$ from Proposition 1 to one, then \eqref{eq:multiplicative_update_expectation} provides a non-increasing update to the multiplicative objective function \eqref{eq:multi}.

%
%
Lastly, we note that the analysis of \citet{cranko_boosting_2019} highlights important properties of the broader class of boosted density estimation models that optimize \eqref{eq:exact_gbnf_objective}, of which both the additive and multiplicative forms of GBNF are members. Specifically, Remark 3 in \citet{cranko_boosting_2019} shows a sharper decrease in the loss---that is, for any step size $\rho \in [0,1]$ the loss has geometric convergence:
\begin{align}
    KL\left( p^* \, || \, \mathbf{G}_{K}^{(c)} | \rho_c \right) \leq (1 - \rho_c) KL\left( p^* \, || \, \mathbf{G}_{K}^{(c-1)} \right)
\end{align}
where $\mathbf{G}_{K}^{(c)} | \rho_c$ denotes the explicit dependence of $\mathbf{G}_{K}^{(c)}(\x)$ on $\rho_c$. Thus GBNF provides a strong convergence guarantee on the global objective.

\subsection{Update to Component Weights}
\label{sec:de_component_weights}
Component weights $\rho$ are updated to satisfy $\rho_c = \argmin_\rho \mathcal{F}^{(ML)}(\phib)$ using line-search.  Alternatively, taking the gradient of the loss $\mathcal{F}_{\phib}^{(ML)}(\x)$ with respect to $\rho_c$ gives a stochastic gradient descent (SGD) algorithm (see Section \ref{sec:component_weights_derivation}, for example).

Updating a component's weight is only needed once after each component converges. We find, however, that results improve by ``fine-tuning'' each component and their weights with additional training after the initial training pass. During the fine-tuning stage, we sequentially retrain each component $g_{K}^{(i)}$ for $i=1, \dots, c$, during which we treat $G_{K}^{(-i)}$ as fixed where $-i$ represents the mixture of all other components: $1, \dots, i-1, i+1, \dots c$. Figure \ref{fig:gbnf_toy} demonstrates this phenomenon: when a single flow is not flexible enough to model the target, mode-covering behavior arises. Introducing the second component trained with the boosting objective improves results, and consequently the second component's weight is increased. Fine-tuning the first component leads to a better solution and assigns equal weight to the two components.

\section{Variational Inference with GBNF}
\label{sec:gbnf_approximate}

Gradient boosting is also applicable to posterior approximation with flow-based models. For variational inference we map a simple base distribution to a complex posterior. Unlike \eqref{eq:nf_elbo}, however, we consider a VAE whose approximate posterior $G_{K}^{(c)}$ is a GBNF with $c$ components and of the form:
\begin{equation}
    G_{K}^{(c)}(\z \mid \x) = (1-\rho_c) G_{K}^{(c-1)}(\z \mid \x) + \rho_c g_{K}^{(c)}(\z \mid  \x)~.
\label{eq:gbnf_vi_model}
\end{equation}
We seek a variational posterior that closely matches the true posterior $p(\z \mid \x)$, which corresponds to the reverse KL-divergence $KL( G_{K}^{(c)}(\z \mid \x) \, || \, p(\z \mid \x) )$. Minimizing KL is equivalent to minimizing $\mathcal{F}_{\phi, \theta}^{(VI)}(\x)$ the negative-ELBO up to a constant. Thus, we seek to minimize the variational bound:
\begin{equation}
\label{eq:gbnf_loss}
\mathcal{F}_{\phi, \theta}^{(VI)}(\x) = \E_{G_{K}^{(c)}} \left[ \log G_{K}^{(c)}(\z_K \mid \x) - \log p_{\theta}(\x, \z_K) \right] ~. \\
\end{equation}

\subsection{Updates to New Boosting Components}
Given the bound  \eqref{eq:gbnf_loss}, we then derive updates for new components. Similar to Section \ref{sec:gbnf_de_update}, consider the functional gradient w.r.t. $G_{K}^{(c)}$ at $\rho_c \rightarrow 0$:
\begin{equation}
\label{eq:gradient}
\nabla_{G_{K}^{(c)}} \mathcal{F}_{\phi, \theta}^{(VI)}(\x) \big|_{\rho_c \rightarrow 0} = - \log \frac{p_\theta(\x , \z)}{G_{K}^{(c-1)}(\z \mid \x)}
\end{equation}
We minimize $\mathcal{F}_{\theta, \phi}^{(VI)}(\x)$ by choosing a new component $g_{K}^{(c)}$ that has the minimum inner product with the gradient from \eqref{eq:gradient}.
\begin{align*}
g_{K}^{(c)} &= \argmin_{g_{K} \in \cG_{K}} ~\sum_{i=1}^n \EE_{g_{K}(\z \mid \x_i)}   \left[ \nabla_G \mathcal{F}(\x_i) \right] \\
\end{align*}
However, to avoid $g_{K}^{(c)}$ degenerating to a point mass at the functional gradient's minimum, we add an entropy regularization term\footnote{In our experiments that augment the VAE with a GBF-based posterior, we find good results setting the regularization $\lambda=1.0$. In the density estimation experiments, better results are often achieved with $\lambda$ near 0.8.} controlled by $\lambda > 0$, thus:
\begin{equation}
\label{eq:g_entropy}
g_{K}^{(c)} = \argmin_{g_{K} \in \cG_{K}} \sum_{i=1}^n \EE_{g_{K}(\z \mid \x_i)} \left[ \nabla_G \mathcal{F}(\x_i) + \lambda \log g_{K}(\z \mid \x_i) \right].
\end{equation}
Despite the differences in derivation, optimization of GBNF has a similar structure to other flow-based VAEs. Specifically, with the addition of the entropy regularization, rearranging \eqref{eq:g_entropy} shows:
\begin{equation}
\label{eq:gbnf}
g_{K}^{(c)} = \argmin_{g_{K} \in \cG_{K}} \EE_{g_{K}(\z \mid \x)} \left[ -\log \frac{ p_\theta(\x \mid \z_K^{(c)}) }{ G_{K}^{(c-1)}(\z_K^{(c)} \mid \x) } \right] +  KL\left( \lambda g_{K}(\z_K^{(c)} \mid \x) \, || \, p(\z_K^{(c)})  \right) ~.
\end{equation}
If $G_K^{(c-1)}$ is constant, then we recover the VAE objective exactly. By learning a GBNF approximate posterior the reconstruction error --$\log p_\theta(\x \mid \z_K^{(c)})$ is down-weighted for samples that are easily explained by the fixed components. For updates to the component weights $\rho$ see Appendix \ref{sec:component_weights_derivation}. 

Lastly, we note that during a forward pass the model encodes data to produce $\z_0$. To sample from the posterior $\z_K \sim G_{K}^{(c)}$, however, we transform $\z_0$ according to $\z_K = f_K^{(j)} \circ \dots \circ f_1^{(j)}(\z_0)$, where $j \sim Categorical(\rho)$ randomly chooses a component---similar to sampling from a mixture model. Thus, during training we compute a fast stochastic approximation of the likelihood $G_K^{(c)}$. Likewise, prediction and sampling are as fast as the non-boosted setting, and easily parallelizable across components.

\subsection{Updating Component Weights}
\label{sec:component_weights_derivation}

\begin{algorithm*}
\caption{Updating Mixture Weight $\rho_c$.}
\label{algo:rhoc}
\DontPrintSemicolon
Let: Tolerance $\epsilon > 0$, and Step-size $\delta > 0$ \;

Initialize weight $\rho_c^{(0)} = 1 / C$ \;

Set iteration $t=0$ \;

\While{$|\rho_c^{(t)} - \rho_c^{(t-1)}| < \epsilon$}{
  Draw mini-batch samples $\z_{K,i}^{(c-1)} \sim G_{K}^{(c-1)}(\z \mid \x_i)$ and $\z_{K,i}^{(c)} \sim g_{K}^{(c)}(\z \mid  \x_i)$ for $i = 1, \dots, n$ \;
  
  Compute Monte Carlo estimate of gradient $\nabla_{\rho_c} \mathcal{F}_{\theta, \phi}^{(VI)}(\x) = \frac{1}{n} \sum_{i=1}^n \gamma_{\rho_c}^{(t-1)}(\z_{K,i}^{(c)} \mid \x_i) - \gamma_{\rho_c}^{(t-1)}(\z_{K,i}^{(c-1)} \mid \x_i)$ \;
  
  t = t + 1 \;
  
  $\rho_c^{(t)} = \rho_c^{(t-1)} - \delta \nabla_{\rho_c}$ \;
  
  $\rho_c^{(t)} = $ clip$(\rho_c^{(t)}, [0, 1])$ \;
}
\Return $\rho_c^{(t)}$ \;
\end{algorithm*}

After $g_{K}^{(c)}(\z_K \mid \x)$ has been estimated, the mixture model still needs to estimate $\rho_c \in [0,1]$. Similar to the density estimation setting, the weights on each component can be updated by taking the gradient of the loss $\mathcal{F}_{\phi, \theta}^{(VI)}(\x)$ with respect to $\rho_c$. Recall that $G_{K}^{(c)}(\z_K \mid \x)$ can be written as the convex combination:
\begin{align*}
G_{K}^{(c)}(\z_K \mid \x) =& (1-\rho_c) G_{K}^{(c-1)}(\z_K \mid \x) + \rho_c g_{K}^{(c)}(\z_K \mid  \x) \\
&= \rho_c \left( g_{K}^{(c)}(\z_K \mid  \x) - G_{K}^{(c-1)}(\z_K \mid \x) \right) + G_{K}^{(c-1)}(\z_K \mid \x)~,
\end{align*}
Then, with $\Delta_{K}^{(c)}(\z_K \mid \x)  \triangleq  g_{K}^{(c)}(\z_K \mid  \x_i) - G_{K}^{(c-1)}(\z_K \mid \x_i)$, the objective function $\mathcal{F}_{\theta, \phi}^{(VI)}(\x)$ can be written as a function of $\rho_c$: 
\begin{align}
\label{eq:rho_objective}
\mathcal{F}_{\theta, \phi}^{(VI)}(\x) &= \sum_{i=1}^n \left\langle \rho_c  \Delta_{K}^{(c)} (\z_K \mid \x_i) + G_{K}^{(c-1)}(\z_K \mid \x_i), - \log p_\theta(\x_i, \z_K) \right\rangle \nonumber \\
&+ \sum_{i=1}^n \bigg\langle \rho_c  \Delta_{K}^{(c)}(\z_K \mid \x_i) + G_{K}^{(c-1)}(\z_K \mid \x_i), \log \left( \rho_c  \Delta_{K}^{(c)}(\z_K \mid \x_i)  + G_{K}^{(c-1)}(\z_K \mid \x_i) \right) \bigg \rangle~.
\end{align}
The above expression can be used in a black-box line search method or, as we have done, in a stochastic gradient descent algorithm \ref{algo:rhoc}. Toward that end, taking gradient of \eqref{eq:rho_objective} w.r.t. $\rho_c$ yields the component weight updates:
\begin{align}
\label{eq:gradient_rhoc}
\frac{ \partial \mathcal{F}_{\phi, \theta}^{(VI)} }{ \partial \rho_c } = \sum_{i=1}^n & \left( \EE_{g_{K}^{(c)}(\z \mid  \x_i)} \left[ \gamma_{\rho_c}^{(t-1)}(\z \mid \x_i) \right] -\EE_{G_{K}^{(c-1)}(\z \mid \x_i)} \left[ \gamma_{\rho_c}^{(t-1)}(\z \mid \x_i) \right] \right)~,
\end{align}
where we've defined:
\begin{align*}
\gamma_{\rho_c}^{(t-1)}(\z \mid \x_i) \triangleq \log \left( \frac{ (1-\rho_c^{(t-1)}) G_{K}^{(c-1)}(\z \mid \x_i) + \rho_c^{(t-1)} g_{K}^{(c)}(\z \mid  \x_i) }{p_\theta(\x_i, \z) } \right)~.
\end{align*}
To ensure a stable convergence we follow \citet{guo_boosting_2016} and implement an SGD algorithm with a decaying learning rate.

\begin{figure}[ht]   
\centering
\includegraphics[width=0.8\textwidth]{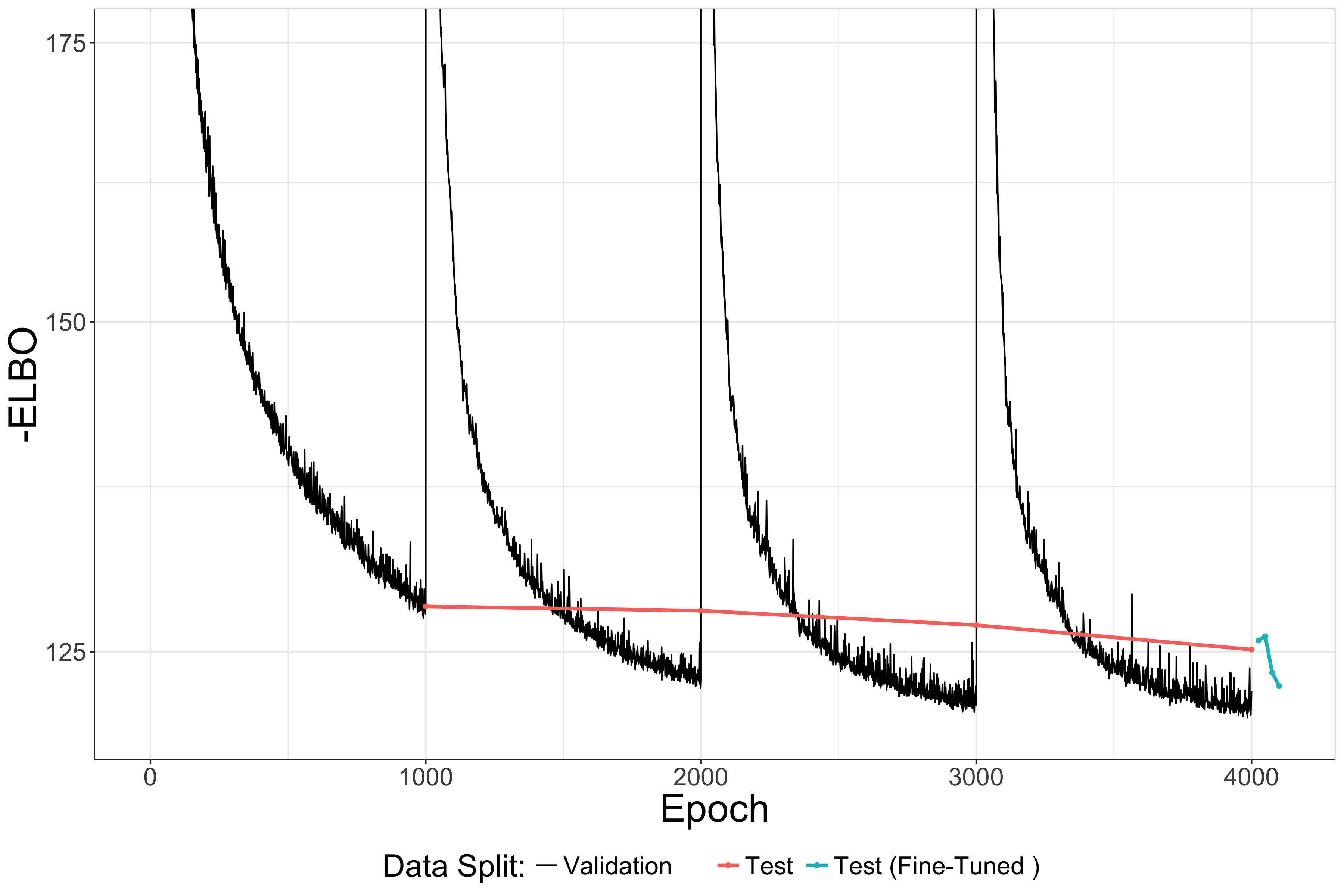}
\caption{``Decoder Shock'' on the Caltech 101 Silhouettes dataset, where new boosting components are introduced every 1000 epochs. While loss on the test set decreases steadily as we add new components, the validation loss jumps dramatically when a new component is introduced due to sudden changes in the distribution of samples passed to the decoder. We also highlight how \emph{fine-tuning} components---by making a second pass with only 25 epochs over each component, improves results at very little computational cost.}
\label{fig:decoder_shock}
\end{figure}

\subsection{Decoder Shock: Abrupt Changes to the VAE Approximate Posterior}

Sharing the decoder between all GBNF components presents a unique challenge in training a VAE with a GBNF approximate posterior. During training the decoder acclimates to samples from a particular component (e.g. $g^{(old)}$). However, when a new stage begins the decoder begins receiving samples from a new component $g^{(new)}$. At this point the loss jumps (see Figure \ref{fig:decoder_shock}), a phenomenon we refer to as ``decoder shock''. Reasons for ``decoder shock'' are as follows.

The introduction of $g^{(new)}$ causes a sudden shift in the distribution of samples passed to the decoder, causing a sharp increase in reconstruction errors. Further, we anneal the KL \citep{bowman_generating_2016,sonderby_ladder_2016,higgins_vvae_2017} in \eqref{eq:gbnf} cyclically \citep{fu_cyclical_2019}, with restarts corresponding to the introduction of new boosting components. By reducing the weight of the KL term in \eqref{eq:gbnf} during the initial epochs the model is free to discover useful representations of the data before being penalized for complexity. Without KL-annealing, models may choose the ``low hanging fruit'' of ignoring $\z$ and relying purely on a powerful decoder \citep{bowman_generating_2016,sonderby_ladder_2016,chen_variational_2017,rainforth_tighter_2018,cremer_inference_2018,higgins_vvae_2017}. Thus, when the annealing schedule restarts, $g^{(new)}$ is unrestricted and the validation's KL term temporarily increases.

A spike in loss between boosting stages is unique to GBNF. Unlike other boosted models, with GBNF there is a module (the decoder) that depends on the boosted components---this does not exist when boosting decision trees for regression or classification (for example). To overcome the ``decoder shock'' problem, propose a simple solution that deviates from a traditional boosting approach. Instead of only drawing samples from $g_{K}^{(c)}$ during training, we periodically sample from the fixed components, helping the decoder \emph{remember} past components and adjust to changes in the full approximate posterior $G_{K}^{(c)}$. We emphasize that despite drawing samples from $G_{K}^{(c-1)}$, the parameters for $G_{K}^{(c-1)}$ remain fixed---samples from $G_{K}^{(c-1)}$ are purely for the decoder's benefit. Additionally, Figure \ref{fig:decoder_shock} highlights how \emph{fine-tuning} (blue line) consolidates information from all components and improves results at very little computational cost.

\section{Related Work}
\label{sec:discussion}

Below we highlight connections between GBNF and related work, along with unique aspects of GBNF. First, we discuss the catalog of normalizing flows that are compatible with gradient boosting. We then compare GBNF to other boosted generative models and flows with mixture formulations.

\begin{figure}
\includegraphics[width=1.0\linewidth]{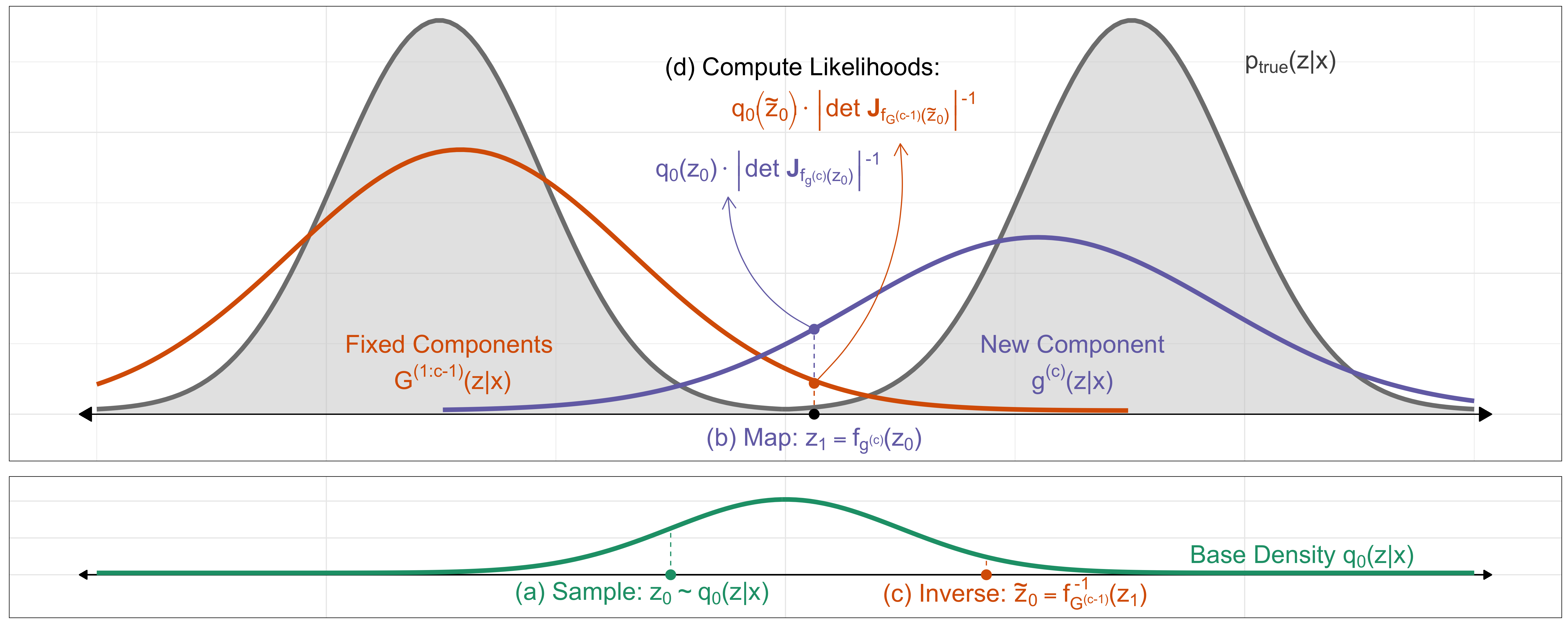}
\caption{Gradient boosted normalizing flows for variational inference require analytically invertible flows. Similar to a traditional flow-based model: \textbf{(a)} samples are drawn from the base density $\z_0 \sim q_0$, and \textbf{(b)} transformed by the $K$-step flow transformation. For GBNF, the sample is transformed by the new component giving $\z_1 = f_{g^{(c)}}(\z_0)$. Gradient boosting fits the new component to the \emph{residuals} of the fixed components, and hence requires computing $G^{(c-1)}(\z_1 \mid \x)$. Due to the change of variables formula, $G^{(c-1)}(\z_1 \mid \x)$ is computed by \textbf{(c)} mapping $\z_1$ back to the base density using the inverse flow transformation  $\widetilde{\z}_0 = f_{G^{(c-1)}}^{-1}(\z_1)$, and then \textbf{(d)} evaluating $q_0( \widetilde{\z}_0 ) \cdot \vert \det \, \mathbf{J}_{f_{G^{(c-1)}}} \vert^{-1}$.}
\label{fig:invertibility}
\end{figure}

\paragraph{Flows Compatible with Gradient Boosting}
While all normalizing flows can be boosted for density estimation, boosting for variational inference is only practical with \emph{analytically} invertible flows (see Figure \ref{fig:invertibility}). The focus of GBNF for variational inference is on training the new component $g_{K}^{(c)}$, but in order to draw samples $\z_K^{(c)} \sim g_{K}^{(c)}$ we sample from the base distribution $\z_0 \sim q_0(\z \mid \x)$ and transform $\z_0$ according to:
\begin{equation*}
\z_K^{(c)} = f_{c,K} \circ \dots \circ f_{c,2} \circ f_{c,1}(\z_0) ~.
\end{equation*}
However, updating $g_{K}^{(c)}$ for variational inference requires computing the likelihood $G_{K}^{(c-1)}(\z_K^{(c)} \mid \x)$. Following Figure \ref{fig:invertibility}, to compute $G_{K}^{(c-1)}$ we seek the point $\widetilde{\z}_0$ within the base distribution such that:
\begin{equation*}
\z_K^{(c)} = f_K^{(j)} \circ \dots \circ f_2^{(j)} \circ f_1^{(j)}(\widetilde{\z}_0) ~,
\end{equation*}
where $\z_K^{(c)}$ is sampled from $g^{(c)}$ and $j \sim Categorical(\rho_{1:c-1})$ randomly chooses one of the fixed components. Then, under the change of variables formula, we approximate $G_{K}^{(c-1)}(\z_K^{(c)} \mid \x)$ by:
\begin{equation*}
q_0(\widetilde{\z}_0) \prod_{k=1}^K \left| \det \frac{\partial f_k^{(j)}}{\partial \widetilde{\z}_{k-1}} \right|^{-1}.
\end{equation*}

While planar and radial \citep{rezende_variational_2015}, Sylvester \citep{vandenberg_sylvester_2018}, and neural autoregressive flows \citep{huang_neural_2018,decao_block_2019} are provably invertible, we cannot compute the inverse. Inverse and masked autoregressive flows \citep{kingma_improving_2016,papamakarios_masked_2017} are invertible, but $D$ times slower to invert where $D$ is the dimensionality of $\z$.

Analytically invertible flows include those based on coupling layers, such as NICE \citep{dinh_nice_2015}, RealNVP \citep{dinh_density_2017}, and Glow---which replaced RealNVP's permutation operation with a $1 \times 1$ convolution \citep{kingma_glow_2018}. Neural spline flows increase the flexibility of both coupling and autoregressive transforms using monotonic rational-quadratic splines \citep{durkan_neural_2019}, and non-linear squared flows \citep{ziegler_latent_2019} are highly multi-modal and can be inverted for boosting. Continuous-time flows \citep{chen_continuoustime_2017,chen_neural_2018,grathwohl_ffjord_2019,salman_deep_2018} use transformations described by ordinary differential equations, with FFJORD being ``one-pass'' invertible by solving an ODE.

\paragraph{Flows with Mixture Formulations}
The main bottleneck in creating more expressive flows lies in the base distribution and the class of transformation function \citep{papamakarios_normalizing_2019}. Autoregressive \citep{huang_neural_2018,jaini_sumofsquares_2019,decao_block_2019,kingma_improving_2016,papamakarios_masked_2017,ma_macow_2019}, residual \citep{behrmann_invertible_2019,chen_residual_2019,vandenberg_sylvester_2018,rezende_variational_2015}, and coupling-layer flows \citep{dinh_nice_2015,dinh_density_2017,kingma_glow_2018,ho_flow_2019,prenger_waveglow_2018} are the most common classes of finite transformations, however, discrete (RAD, \citep{dinh_rad_2019}) and continuous (CIF, \citep{cornish_relaxing_2020}) mixture formulations offer a promising new approach where the base distribution and transformation change according to the mixture component. GBNF also presents a mixture formulation, but trained in a different way, where only the updates to the newest component are needed during training and extending an existing model with additional components is trivial. Moreover, GBNF optimizes a different objective that fits new components to the residuals of previously trained components, which can refine the \emph{mode covering} behavior of VAEs (see \citet{hu_unifying_2018}) and maximum likelihood (similar to \citet{dinh_rad_2019}). The continuous mixture approach of CIF, however, cannot be used in the variational inference setting to augment the VAE's approximate posterior \citep{cornish_relaxing_2020}.

\paragraph{Gradient Boosted Generative Models}
By considering convex combinations of distributions $G$ and $g$, boosting is applicable beyond the traditional supervised setting \citep{campbell_universal_2019,cranko_boosting_2019,grover_boosted_2018,guo_boosting_2016,lebanon_boosting_2002,locatello_boosting_2018,rosset_boosting_2002,tolstikhin_adagan_2017}.  In particular, boosting variational inference (BVI, \citep{miller_variational_2017,guo_boosting_2016,cranko_boosting_2019}) improves a variational posterior, and boosted generative models (BGM, \citep{grover_boosted_2018}) constructs a density estimator by iteratively combining sum-product networks. Unlike BVI and BGM our approach addresses the unique algorithmic challenges of boosting applied to flow-based models---such as the need for analytically invertible flows and the ``decoder shock'' phenomena when enhancing the VAE's approximate posterior with GBNF.

\section{Experiments}
\label{sec:experiments}

To evaluate GBNF, we highlight results on two toy problems, density estimation on real data, and boosted flows within a VAE for generative modeling of images. We boost coupling flows \cite{dinh_density_2017,kingma_glow_2018}  parameterized by feed-forward networks with TanH activations and a single hidden layer. While RealNVP \citep{dinh_density_2017}, in particular, is less flexible and shown to be empirically inferior to planar flows in variational inference \citep{rezende_variational_2015}, coupling flows are attractive for boosting: sampling and inference require one forward pass, log-likelihoods are computed exactly, and they are trivially invertible. In the toy experiments flows are trained for 25k iterations using the Adam optimizer \citep{kingma_adam_2015}. For all other experiments details on the datasets and hyperparameters can be found in Appendix \ref{sec:datasets}.

%
%
\newcommand{\matchingfigwidth}{0.14\linewidth}
\begin{figure}
\centering

\begin{subfigure}[b]{3mm}
\centering
1
\vspace{2em}
\end{subfigure}%
\begin{subfigure}[b]{\matchingfigwidth}
\centering
\caption*{Target}
\vspace{-2mm}
\includegraphics[width=\linewidth]{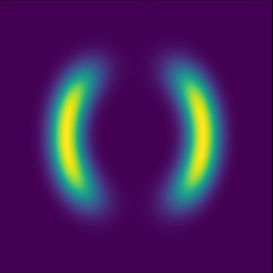}
\end{subfigure}%
\hspace{1mm}%
\begin{subfigure}[b]{\matchingfigwidth}
\centering
\caption*{\centering RealNVP}
\vspace{-2mm}
\includegraphics[width=\linewidth]{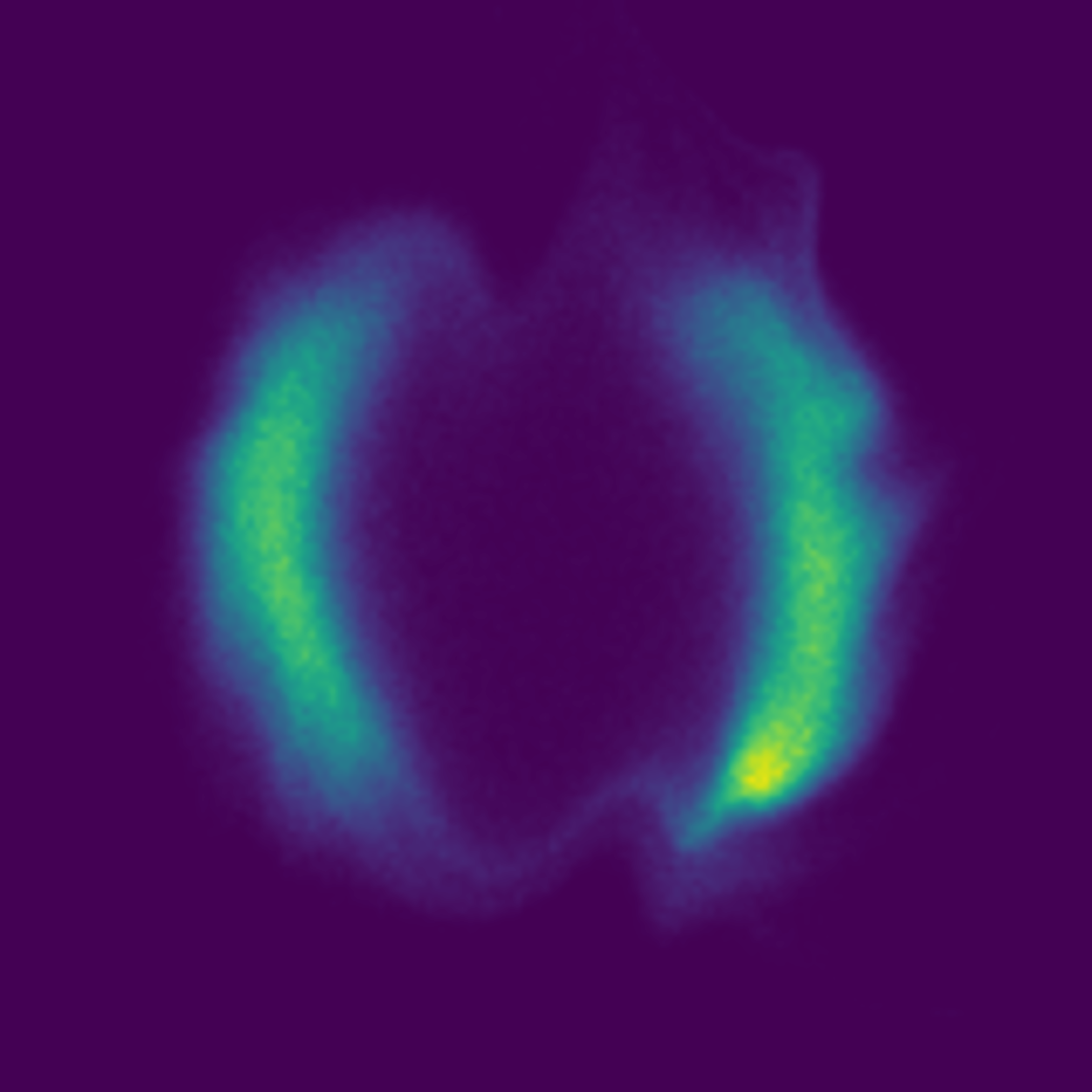}
\end{subfigure}%
\hspace{1mm}%
\begin{subfigure}[b]{\matchingfigwidth}
\centering
\caption*{\centering \textbf{GBNF}}
\vspace{-2mm}
\includegraphics[width=\linewidth]{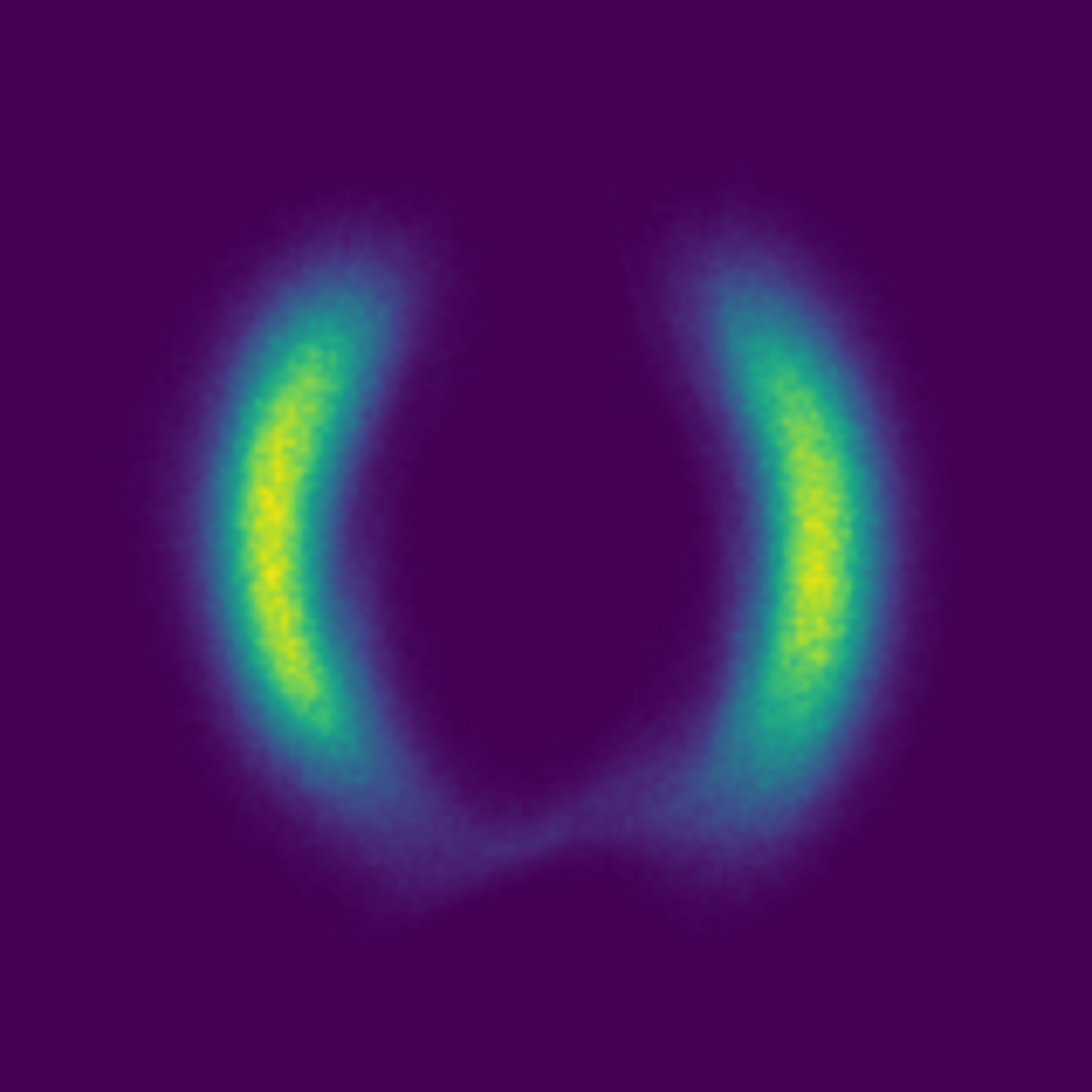}
\end{subfigure}%
\hspace{6mm}%
\begin{subfigure}[b]{3mm}
\centering
3
\vspace{2em}
\end{subfigure}%
\begin{subfigure}[b]{\matchingfigwidth}
\centering
\caption*{Target}
\vspace{-2mm}
\includegraphics[width=\linewidth]{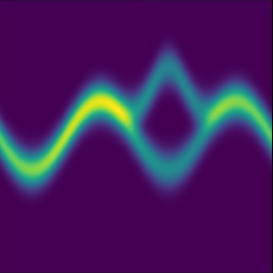}
\end{subfigure}%
\hspace{1mm}%
\begin{subfigure}[b]{\matchingfigwidth}
\centering
\caption*{\centering RealNVP}
\vspace{-2mm}
\includegraphics[width=\linewidth]{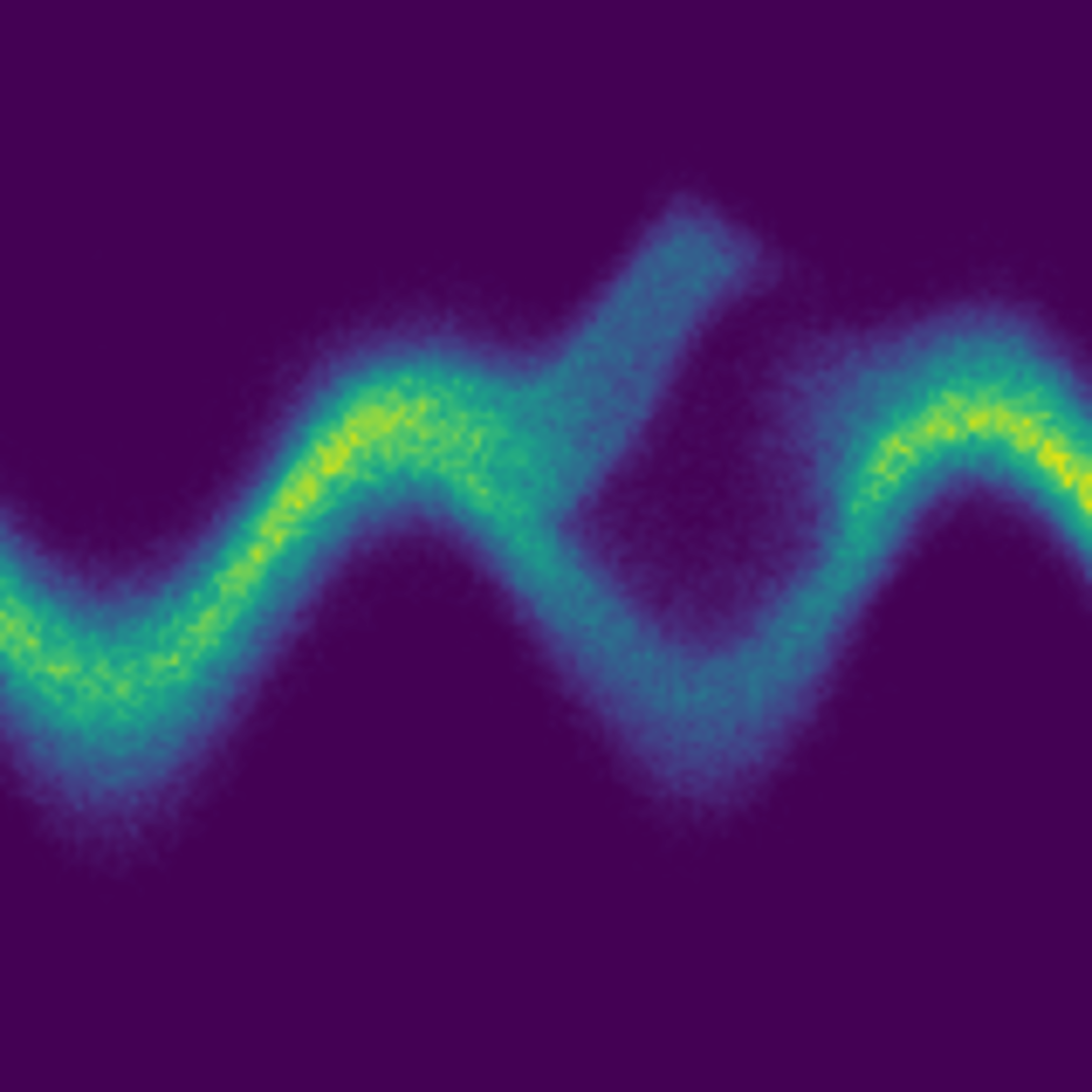}
\end{subfigure}%
\hspace{1mm}%
\begin{subfigure}[b]{\matchingfigwidth}
\centering
\caption*{\centering \textbf{GBNF}}
\vspace{-2mm}
\includegraphics[width=\linewidth]{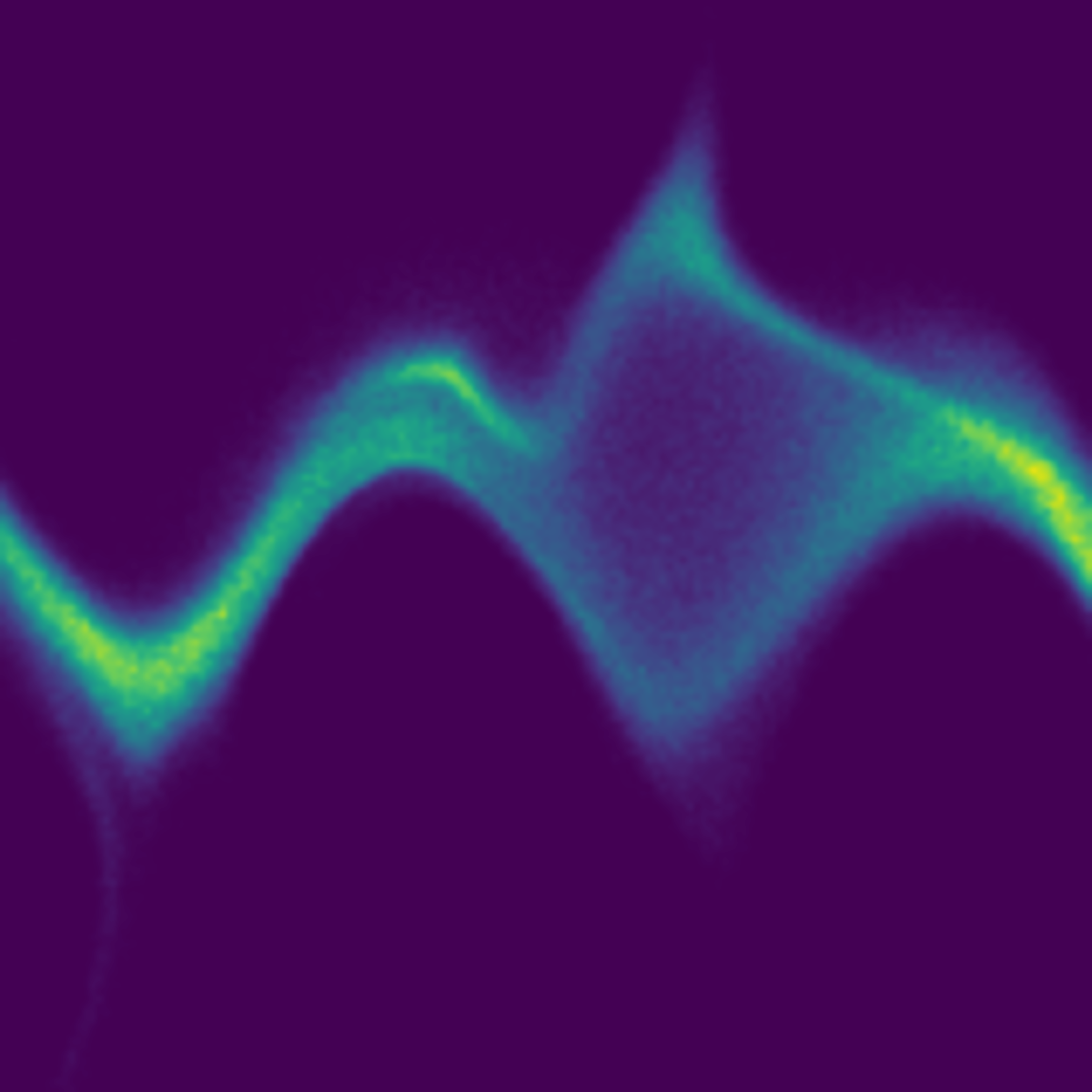}
\end{subfigure}%

\vspace{1em}
\begin{subfigure}[b]{3mm}
\centering
2
\vspace{2em}
\end{subfigure}%
\begin{subfigure}[b]{\matchingfigwidth}
\centering
\vspace{-2mm}
\includegraphics[width=\linewidth]{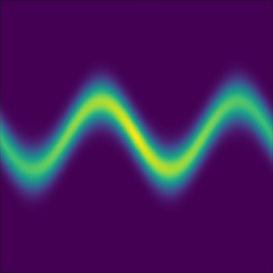}
\end{subfigure}%
\hspace{1mm}%
\begin{subfigure}[b]{\matchingfigwidth}
\centering
\vspace{-2mm}
\includegraphics[width=\linewidth]{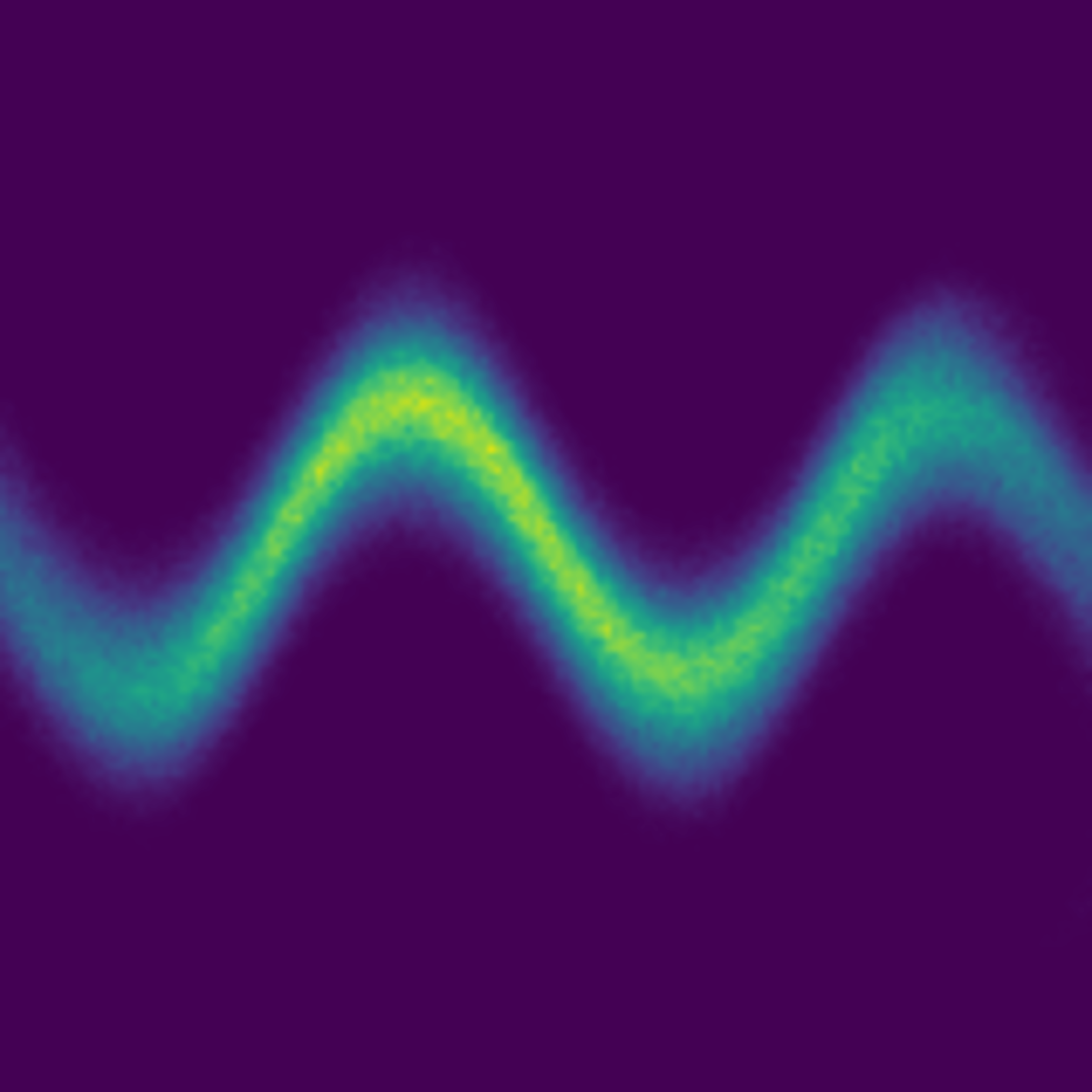}
\end{subfigure}%
\hspace{1mm}%
\begin{subfigure}[b]{\matchingfigwidth}
\centering
\vspace{-2mm}
\includegraphics[width=\linewidth]{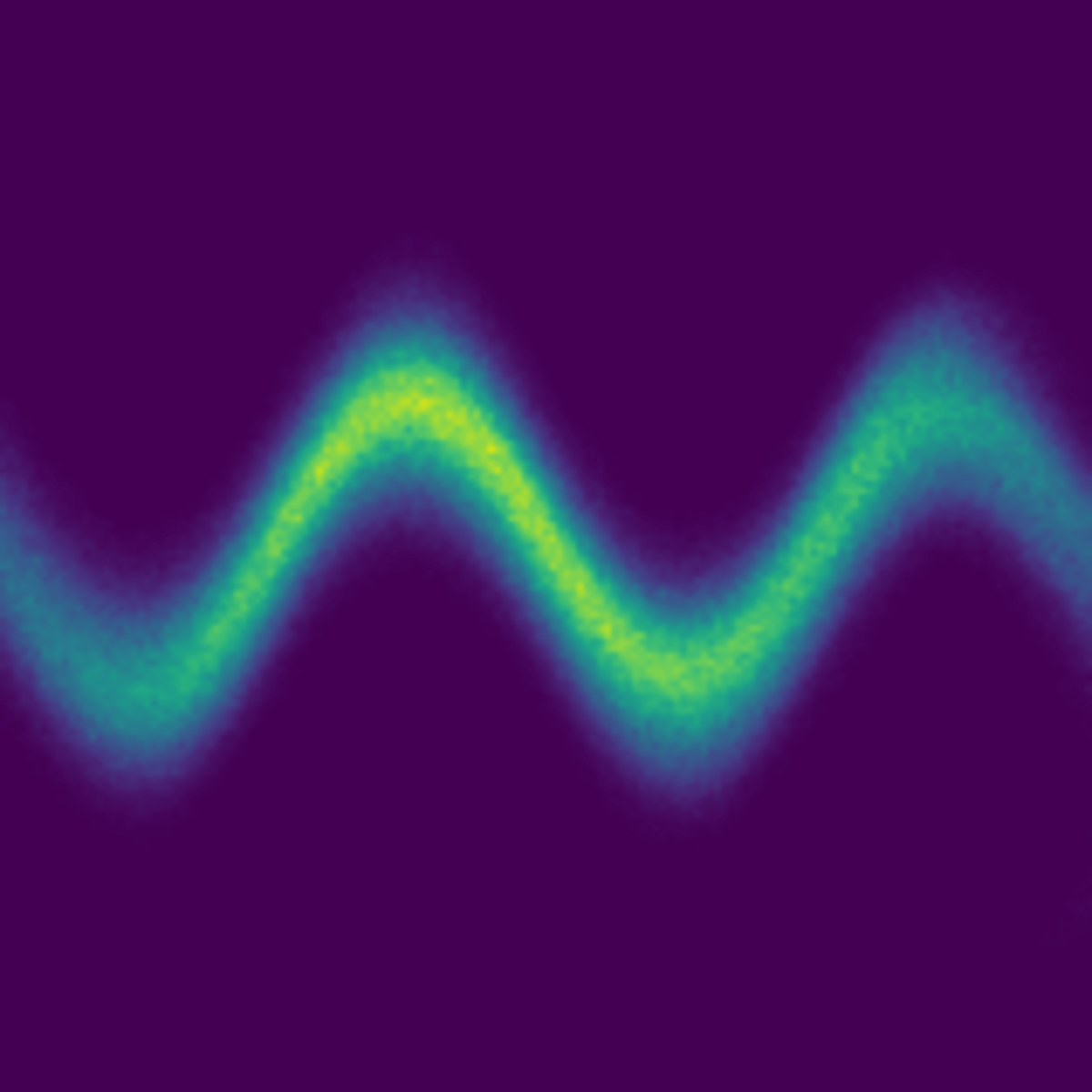}
\end{subfigure}%
\hspace{6mm}%
\begin{subfigure}[b]{3mm}
\centering
4
\vspace{2em}
\end{subfigure}%
\begin{subfigure}[b]{\matchingfigwidth}
\centering
\vspace{-2mm}
\includegraphics[width=\linewidth]{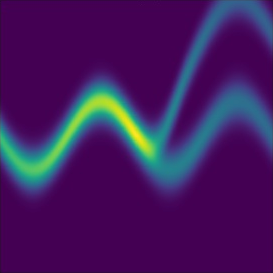}
\end{subfigure}%
\hspace{1mm}%
\begin{subfigure}[b]{\matchingfigwidth}
\centering
\vspace{-2mm}
\includegraphics[width=\linewidth]{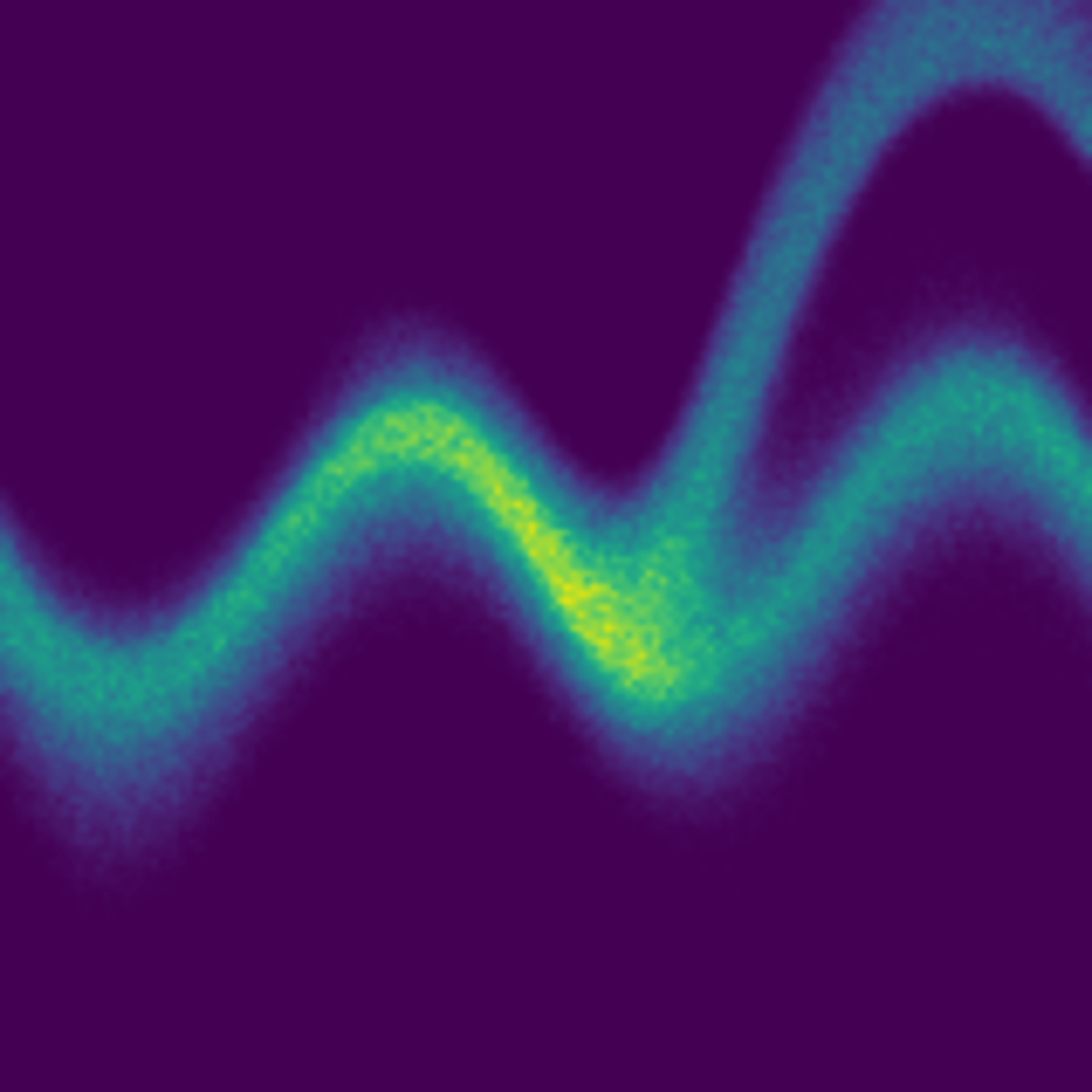}
\end{subfigure}%
\hspace{1mm}%
\begin{subfigure}[b]{\matchingfigwidth}
\centering
\vspace{-2mm}
\includegraphics[width=\linewidth]{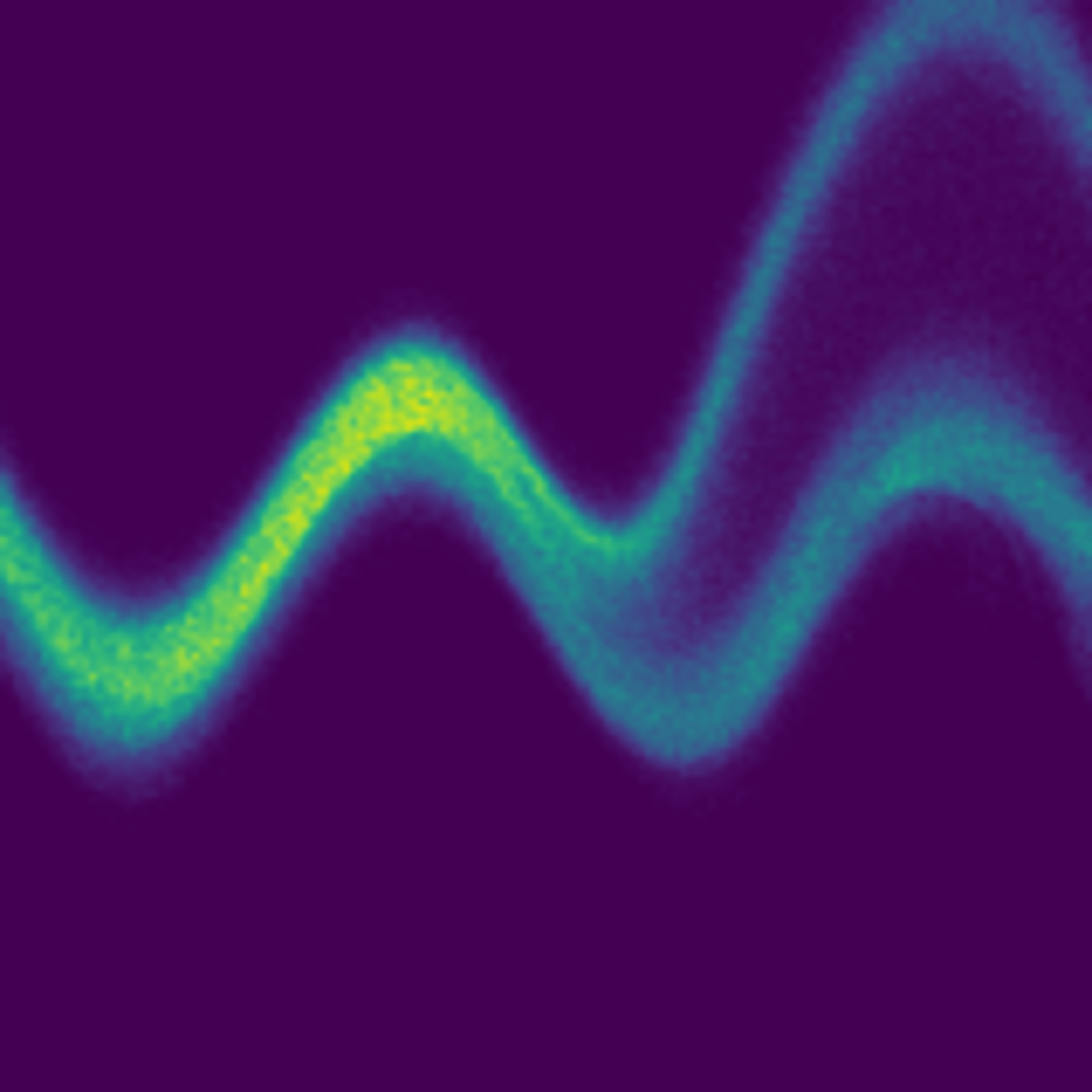}
\end{subfigure}%
\caption{Matching the energy functions from Table 1 of \citet{rezende_variational_2015}. The middle columns show deep RealNVPs with $K=16$ flows. Gradient boosting RealNVP with $c=2$ components of length $K=4$ performs as well or better with \textbf{half as many parameters}.}
\label{fig:density_matching}
\end{figure}

\subsection{Toy Density Matching}
For density matching the model generates samples from a standard normal and transforms them into a complex distribution $p_{X}$. The 2-dimensional unnormalized target's analytical form $p^*$ is known and parameters are learned by minimizing $KL(p_{X} \, || \, p^*)$.

\paragraph{Results} In Figure \ref{fig:density_matching} we compare our results to a deep 16-step RealNVP flow on four energy functions. In each case GBNF provides an accurate density estimation with half as many parameters. When the component flows are flexible enough to model most or all of the target density, components can overlap. However, by training the component weights $\rho$ the model down-weights  components that don't provide additional information. On more challenging targets, such as \texttt{3} (top-right), GBNF fits one component to each of the top and bottom divergences within the energy function, and some component overlap occurring elsewhere.

\subsection{Toy Density Estimation}
\label{sec:main_toy_density_estimation}

%
%
\newcommand{\figwidth}{0.14\linewidth}
\begin{figure}
\centering

\begin{subfigure}[b]{6mm}
\centering
\caption*{\underline{K}}
\vspace{7mm}
1
\vspace{2em}
\end{subfigure}%
\begin{subfigure}[b]{\figwidth}
\centering
\caption*{Data}
\vspace{-2mm}
\includegraphics[width=\linewidth]{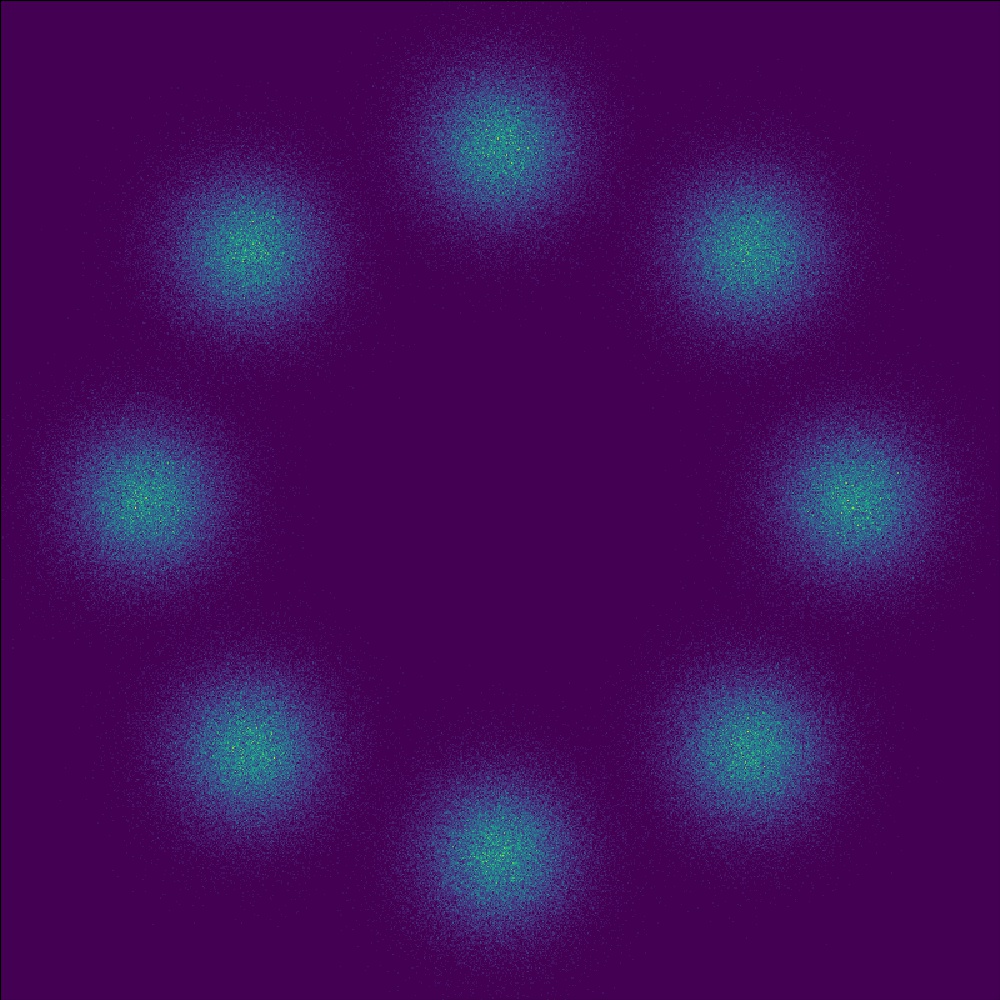}
\end{subfigure}%
\hspace{1mm}%
\begin{subfigure}[b]{\figwidth}
\centering
\caption*{\centering RealNVP}
\vspace{-2mm}
\includegraphics[width=\linewidth]{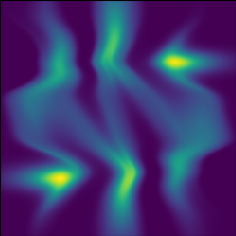}
\end{subfigure}%
\hspace{1mm}%
\begin{subfigure}[b]{\figwidth}
\centering
\caption*{\centering \textbf{GBNF}}
\vspace{-2mm}
\includegraphics[width=\linewidth]{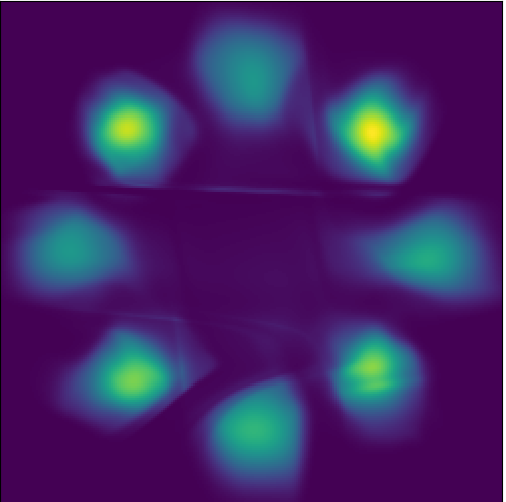}
\end{subfigure}%
\hspace{6mm}%
\begin{subfigure}[b]{6mm}
\centering
\caption*{\underline{K}}
\vspace{7mm}
2
\vspace{2em}
\end{subfigure}%
\begin{subfigure}[b]{\figwidth}
\centering
\caption*{\centering Data}
\vspace{-2mm}
\includegraphics[width=\linewidth]{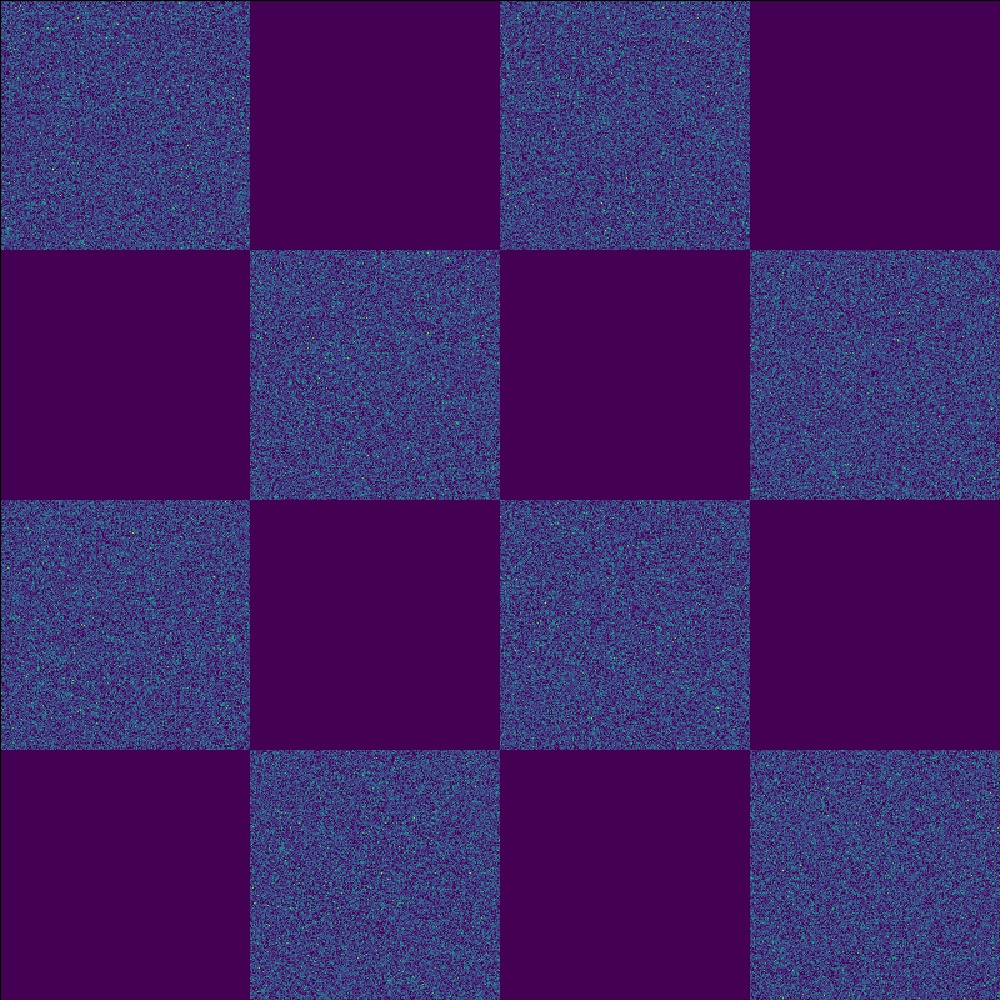}
\end{subfigure}%
\hspace{1mm}%
\begin{subfigure}[b]{\figwidth}
\centering
\caption*{\centering RealNVP}
\vspace{-2mm}
\includegraphics[width=\linewidth]{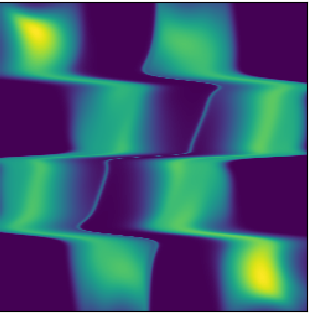}
\end{subfigure}%
\hspace{1mm}%
\begin{subfigure}[b]{\figwidth}
\centering
\caption*{\centering \textbf{GBNF}}
\vspace{-2mm}
\includegraphics[width=\linewidth]{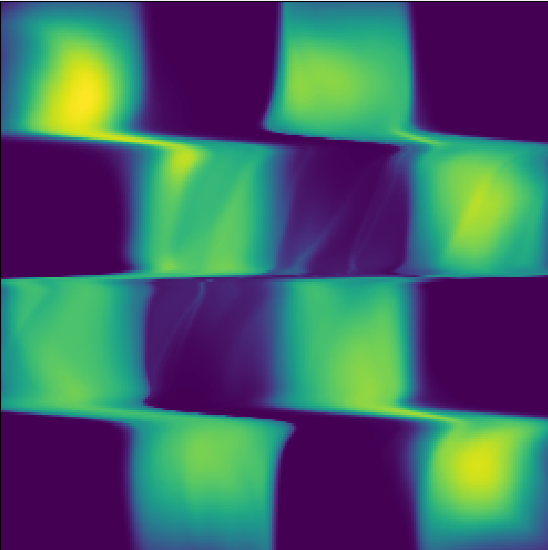}
\end{subfigure}%

\vspace{1em}
\begin{subfigure}[b]{6mm}
\centering
4
\vspace{2em}
\end{subfigure}%
\begin{subfigure}[b]{\figwidth}
\centering
\vspace{-2mm}
\includegraphics[width=\linewidth]{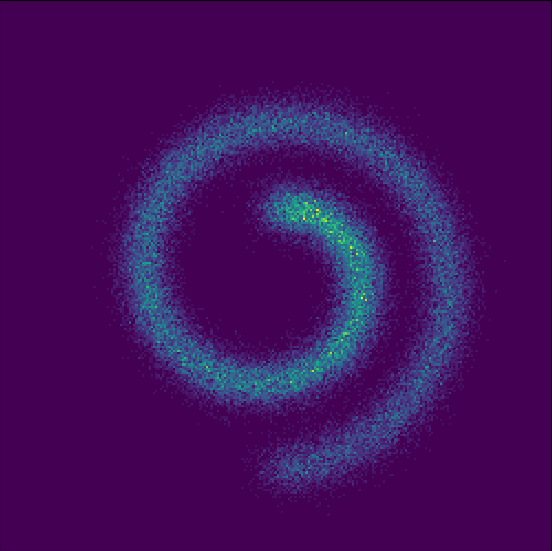}
\end{subfigure}%
\hspace{1mm}%
\begin{subfigure}[b]{\figwidth}
\centering
\vspace{-2mm}
\includegraphics[width=\linewidth]{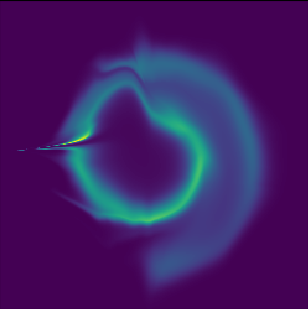}
\end{subfigure}%
\hspace{1mm}%
\begin{subfigure}[b]{\figwidth}
\centering
\vspace{-2mm}
\includegraphics[width=\linewidth]{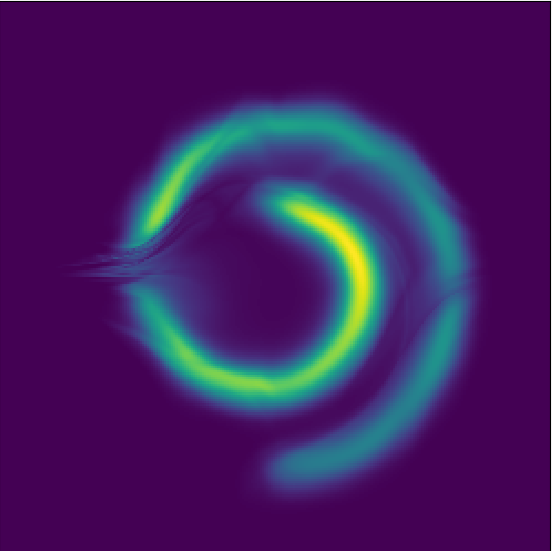}
\end{subfigure}%
\hspace{6mm}%
\begin{subfigure}[b]{6mm}
\centering
8
\vspace{2em}
\end{subfigure}%
\begin{subfigure}[b]{\figwidth}
\centering
\vspace{-2mm}
\includegraphics[width=\linewidth]{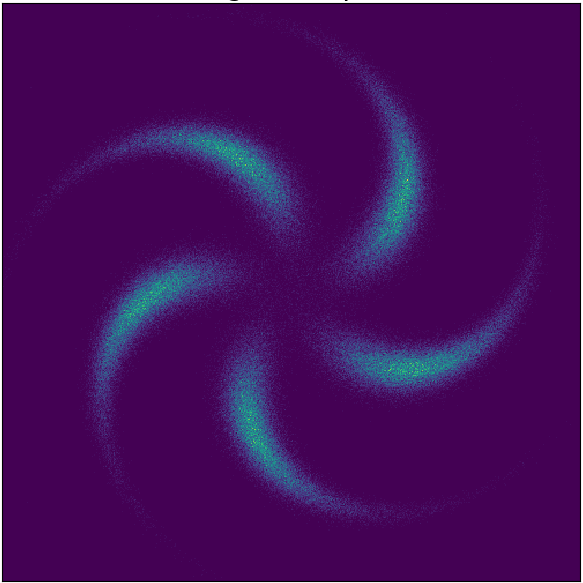}
\end{subfigure}%
\hspace{1mm}%
\begin{subfigure}[b]{\figwidth}
\centering
\vspace{-2mm}
\includegraphics[width=\linewidth]{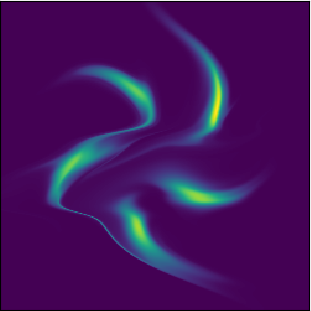}
\end{subfigure}%
\hspace{1mm}%
\begin{subfigure}[b]{\figwidth}
\centering
\vspace{-2mm}
\includegraphics[width=\linewidth]{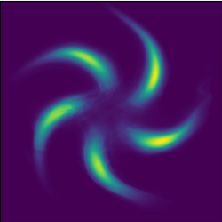}
\end{subfigure}%
\caption{Density estimation for 2D toy data. The \textbf{GBNF} columns shows results for a gradient boosted model where each component is a RealNVP flow with $K=1, 2, 4$ or $8$ flow steps, respectively. For comparison the \textbf{RealNVP} column shows results for a single RealNVP model, and is equivalent to GBNF's first component. GBNF models train $c=4$ components, except on the 8-Gaussians data (top left) where results continued to improve up to 8 components. Results show that GBNF produces more accurate density estimates without increasing the complexity of the flow transformations.}
\label{fig:density_sampling}
\end{figure}

We apply GBNF to the density estimation problems found in \citep{kingma_glow_2018,grathwohl_ffjord_2019,decao_block_2019}. Here the model receives samples from an unknown 2-dimensional data distribution, and the goal is to learn a density estimator of the data. We consider GBNF with either $c=4$ or $8$ RealNVP components, each of which includes $K=1, 2, 4,$ or $8$ coupling layers \citep{dinh_density_2017}, respectively. Here RealNVP and GBNF use flows of equivalent depth, and we evaluate improvements resulting from GBNF's additional boosted components.

\paragraph{Results} As shown in Figure  \ref{fig:density_sampling}, even when individual components are weak the composite model is expressive. For example, the 8-Gaussians figure shows that the first component (RealNVP column) fails to model all modes. With additional 1-step flows, GNBF achieves a multimodal density model. Both the 8-Gaussians and Spiral results show that adding boosted components can drastically improve density estimates without requiring more complex transformations. On the Checkerboard and Pinwheel, where RealNVP matches the data more closely, GBNF sharpens density estimates.

\subsection{Density Estimation on Real Data}

Following \citet{grathwohl_ffjord_2019} we report density estimation results on the POWER, GAS, HEPMASS, and MINIBOONE datasets from the UCI machine learning repository \citep{dua_uci_2017}, as well as the BSDS300 dataset \citep{martin_database_2001}. We compare boosted and non-boosted RealNVP \citep{dinh_density_2017} and Glow models \citep{kingma_glow_2018}. Glow uses a learned base distribution, whereas our boosted implementation of Glow (and the RealNVPs) use fixed Gaussians. Results for non-boosted models are from \citep{grathwohl_ffjord_2019}.

\paragraph{Results} In Table \ref{tab:density} we find significant improvements by boosting Glow and the more simple RealNVP normalizing flows, even with only $c=4$ components. Our implementation of Glow was unable to match the results for BSDS300 from \citep{grathwohl_ffjord_2019}, and only achieves an average log-likelihood of 152.96 without boosting. After boosting Glow with $c=4$ components, however, the log-likelihood rises significantly to 154.68, which is comparable to the baseline.

\begin{table}
\centering
\caption{Log-likelihood on the test set (higher is better) for 4 datasets from UCI machine learning \citep{dua_uci_2017} and BSDS300 \citep{martin_database_2001}. Here $d$ is the dimensionality of data-points and $n$ the size of the dataset. GBNF models include $c=4$ components. Mean/stdev are estimated over 3 runs.}
\begin{tabular}{lccccc}
\toprule
\multirow{2}{*}{\textbf{Model}} & POWER$\uparrow$ & GAS$\uparrow$ & HEPMASS$\uparrow$ & MINIBOONE$\uparrow$ & BSDS300$\uparrow$ \\
 & {\tiny$d$=6;$n$=2,049,280 } &{\tiny $d$=8;$n$=1,052,065} &{\tiny $d$=21;$n$=525,123} &{\tiny$d$=43;$n$=36,488} & {\tiny$d$=63;$n$=1,300,000} \\
\midrule
RealNVP & $0.17${\tiny$\pm .01$} & $8.33${\tiny$\pm .14$} & $-18.71${\tiny$\pm .02$} & $-13.55${\tiny$\pm .49$} & $153.28${\tiny$\pm 1.78$} \\
\textbf{Boosted RealNVP} & $0.27${\tiny$\pm 0.01$} & $9.58${\tiny$\pm .04$} & $-18.60${\tiny$\pm 0.06$} & $-10.69${\tiny$\pm 0.07$} & $154.23${\tiny$\pm 2.21$} \\
\midrule
Glow & $0.17${\tiny$\pm .01$} & $8.15${\tiny$\pm .40$} & $-18.92${\tiny$\pm .08$} & $-11.35${\tiny$\pm .07$} & $155.07${\tiny$\pm .03$} \\
\textbf{Boosted Glow} & $0.24${\tiny$\pm 0.01$} & $9.95${\tiny$\pm 0.11$} & $-17.81${\tiny$\pm 0.12$} & $-10.76${\tiny$\pm 0.02$} & $154.68${\tiny$\pm 0.34$} \\
\bottomrule
\end{tabular}
\label{tab:density}
\end{table}

\subsection{Image Modeling with Variational Autoencoders}
\label{sec:real_results}
Following \citet{rezende_variational_2015}, we employ NFs for improving the VAE's approximate posterior \citep{kingma_autoencoding_2014}. We compare our model on the same image datasets as those used in \citet{vandenberg_sylvester_2018}: Freyfaces, Caltech 101 Silhouettes \citep{marlin_inductive_2010}, Omniglot \citep{lake_humanlevel_2015}, and statically binarized MNIST \citep{larochelle_neural_2011}. 

\paragraph{Results} In Table \ref{tab:vae} we compare the performance of GBNF to other normalizing flow architectures. In all results RealNVP, which is more ideally suited for density estimation tasks, performs the worst of the flow models. Nonetheless, applying gradient boosting to RealNVP improves the results significantly. On Freyfaces, the smallest dataset consisting of just 1965 images, gradient boosted RealNVP gives the best performance---suggesting that GBNF may avoid overfitting. For the larger Omniglot dataset of hand-written characters, Sylvester flows are superior, however, gradient boosting improves the RealNVP baseline considerably and is comparable to Sylvester. GBNF improves on the baseline RealNVP, however both GBNF and IAF's results are notably higher than the non-coupling flows on the Caltech 101 Silhouettes dataset. Lastly, on MNIST we find that boosting improves NLL on RealNVP, and is on par with Sylvester flows. All models have an approximately equal number of parameters, except the baseline VAE (fewer parameters) and Sylvester which has $\approx 5$x the number of parameters (grid search for hyperparameters is chosen following \citep{vandenberg_sylvester_2018}).

%
%
\begin{table}
\caption{Negative ELBO (lower is better) and Negative log-likelihood (NLL, lower is better) results on MNIST, Freyfaces, Omniglot, and Caltech 101 Silhouettes datasets. For the Freyfaces dataset the results are reported in bits per dim. Results for the other datasets are reported in nats. GBNF models include $c=4$ RealNVP components. The top 3 NLL results for each dataset are in \textbf{bold}.}
\resizebox{\linewidth}{!}{%
\begin{tabular}{lcccccccc}
\toprule
\multirow{2}{*}{\bfseries{Model}} & \multicolumn{2}{c}{\bfseries MNIST} & \multicolumn{2}{c}{\bfseries Freyfaces} & \multicolumn{2}{c}{\bfseries Omniglot} & \multicolumn{2}{c}{\bfseries Caltech 101} \\
& \multicolumn{1}{c}{-ELBO$\downarrow$} & \multicolumn{1}{c}{NLL$\downarrow$} & \multicolumn{1}{c}{-ELBO$\downarrow$} & \multicolumn{1}{c}{NLL$\downarrow$} & \multicolumn{1}{c}{-ELBO$\downarrow$} &\multicolumn{1}{c}{NLL$\downarrow$} & \multicolumn{1}{c}{-ELBO$\downarrow$} &\multicolumn{1}{c}{NLL$\downarrow$} \\
\midrule
VAE & $89.32${\tiny$\pm 0.07$} & $84.97${\tiny$\pm 0.01$} & $4.84${\tiny$\pm  0.07$} & $4.78${\tiny$\pm 0.07$} & $109.77${\tiny$\pm 0.06$} & $103.16${\tiny$\pm 0.01$} & $120.98${\tiny$\pm 1.07$} & $108.43${\tiny$\pm 1.81$} \\

Planar & $86.47${\tiny$\pm 0.09$} & $83.16${\tiny$\pm 0.07$} & $4.64${\tiny$\pm 0.04$} & \textbf{4.60{\tiny$\pm 0.04$}} & $105.72${\tiny$\pm 0.08$} & \textbf{100.18{\tiny$\pm 0.01$}} & $116.70${\tiny$\pm 1.70$} & \textbf{104.23 {\tiny$\pm 1.60$}} \\

Radial & $88.43${\tiny$\pm 0.07$} & $84.32${\tiny$\pm 0.06$} & $4.73${\tiny$\pm 0.08$} & $4.68${\tiny$\pm 0.07$} & $108.74${\tiny$\pm 0.57$} & $102.07${\tiny$\pm 0.50$} & $118.89${\tiny$\pm 1.30$} & $106.88${\tiny$\pm 1.55$} \\

Sylvester & 84.54{\tiny$\pm 0.01$} & \textbf{81.99{\tiny$\pm 0.02$}} & 4.54{\tiny$\pm 0.03$} & \textbf{4.49{\tiny$\pm 0.03$}} & 101.99{\tiny$\pm 0.23$} & \textbf{98.54{\tiny$\pm 0.29$}} & 112.26{\tiny$\pm 2.01$} & \textbf{100.38{\tiny$\pm 1.20$}} \\

IAF & $86.46${\tiny$\pm 0.07$} & \textbf{83.14 {\tiny$\pm 0.06$}} & $4.73${\tiny$\pm 0.04$} & $4.70${\tiny$\pm 0.05$} & $106.34${\tiny$\pm 0.14$} & $100.97${\tiny$\pm 0.07$} & $119.62${\tiny$\pm 0.84$} & $108.41${\tiny$\pm 1.31$} \\

RealNVP & $88.04${\tiny$\pm 0.07$} & $83.36${\tiny$\pm 0.09$} & $4.66${\tiny$\pm 0.17$} & $4.62${\tiny$\pm 0.16$} & $106.22${\tiny$\pm 0.59$} & $100.43${\tiny$\pm 0.19$} & $123.26${\tiny$\pm 2.06$} & $113.00${\tiny$\pm 1.70$} \\

\midrule

GBNF & $87.00${\tiny$\pm 0.16$} & \textbf{82.59 {\tiny$\pm 0.03$}} & $4.49${\tiny$\pm 0.01$} & \textbf{4.41 {\tiny$\pm 0.01$}} & $105.60${\tiny$\pm 0.20$} & \textbf{99.09 {\tiny$\pm 0.17$}} & $121.41${\tiny$\pm 0.71$} & \textbf{106.40 {\tiny$\pm 0.54$}} \\
\bottomrule
\end{tabular}%
}
\label{tab:vae}
\end{table}

\section{Conclusion}
\label{sec:conclusion}

In this work we introduce \emph{gradient boosted normalizing flows}, a technique for increasing the flexibility of flow-based models through gradient boosting. GBNF, iteratively adds new NF components, where each new component is fit to the residuals of the previously trained components. We show that GBNF can improve results for existing normalizing flows on density estimation and variational inference tasks. In our experiments we demonstrated that GBNF improves over their baseline single component model, without increasing the depth of the model, and produces image modeling results on par with state-of-the-art flows. Further, we showed GBNF models used for density estimation create more flexible distributions at the cost of additional training and not more complex transformations.

In the future we wish to further investigate the ``decoder shock'' phenomenon occurring when GBNF is paired with a VAE. Future work may benefit from exploring other strategies for alleviating ``decoder shock'', such as multiple decoders or different annealing strategies. In our real data experiments in Section \ref{sec:real_results} we fixed the entropy regularization $\lambda$ at 1.0, however adjusting the regularization on a per-component level may be worth pursuing. Additionally, in our image modeling experiments we used RealNVP as the base component. Future work may consider other flows for boosting, as well as heterogeneous combinations of flows as the different components.
\section{Broader Impact}
\label{sec:broader_impact}

As a generative model, gradient boosted normalizing flows (GBNF) are suited for a variety of tasks, including the synthesis of new data-points. A primary motivation for choosing GBNF, in particular, is producing a flexible model that can synthesize new data-points quickly. GBNF's individual components can be less complex and thus faster, yet as a whole the model is powerful. Since the components operate in parallel, prediction and sampling can be done quickly---a valuable characteristic for deployment on mobile devices. One limitation of GBNF is the requirement for additional computing resources to train the added components, which can be costly for deep flow-based models. As such, GBNF advantages research laboratories and businesses with access to scalable computing. Those with limited computing resources may find benchmarking or deploying GBNF too costly.

\section*{Acknowledgements}
The research was supported by NSF grants OAC-1934634, IIS-1908104, IIS-1563950, IIS-1447566, IIS-1447574, IIS-1422557, CCF-1451986. We thank the University of Minnesota Supercomputing Institute (MSI) for technical support.

\newpage
\bibliography{library}
\bibliographystyle{apalike}

\newpage
\appendix

\section{Experiment Details}
\label{sec:datasets}

\subsection{Image Modeling}

\paragraph{Datasets} In Section \ref{sec:real_results}, VAEs are modified with GBNF approximate posteriors to model four datasets: Freyfaces\footnote{\url{http://www.cs.nyu.edu/~roweis/data/frey_rawface.mat}}, Caltech 101 Silhouettes\footnote{\url{https://people.cs.umass.edu/~marlin/data/caltech101_silhouettes_28_split1.mat}} \citep{marlin_inductive_2010}, Omniglot\footnote{\url{https://github.com/yburda/iwae/tree/master/datasets/OMNIGLOT}} \citep{lake_humanlevel_2015}, and statically binarized MNIST\footnote{\url{http://yann.lecun.com/exdb/mnist/}} \citep{larochelle_neural_2011}. Details of these datasets are given below.

The Freyfaces dataset contains 1965 gray-scale images of size $28 \times 20$ portraying one man's face in a variety of emotional expressions. Following \citet{vandenberg_sylvester_2018}, we randomly split the dataset into 1565 training, 200 validation, and 200 test set images.

The Caltech 101 Silhouettes dataset contains 4100 training, 2264 validation, and 2307 test set images. Each image portrays the black and white silhouette of one of 101 objects, and is of size $28 \times 28$. As \citet{vandenberg_sylvester_2018} note, there is a large variety of objects relative to the training set size, resulting in a particularly difficult modeling challenge.

The Omniglot dataset contains 23000 training, 1345 validation, and 8070 test set images. Each image portrays one of 1623 hand-written characters from 50 different alphabets, and is of size $28 \times 28$. Images in Omniglot are dynamically binarized.

Finally, the MNIST dataset contains 50000 training, 10000 validation, and 10000 test set images. Each $28 \times 28$ image is a binary, and portrays a hand-written digit.

\paragraph{Experimental Setup} We limit the computational complexity of the experiments by reducing the number of convolutional layers in the encoder and decoder of the VAEs from 14 layers to 6. In Table \ref{tab:vae} we compare the performance of our GBNF to other normalizing flow architectures. Planar, radial, and Sylvester normalizing flows each use $K=16$, with Sylvester's bottleneck set to $M=32$ orthogonal vectors per orthogonal matrix. IAF is trained with $K=8$ transformations, each of which is a single hidden layer MADE \citep{germain_made_2015} with either $h=256$ or $512$ hidden units. RealNVP uses $K=8$ transformations with either $h=256$ or $h=512$ hidden units in the Tanh feed-forward network. For all models, the dimensionality of the flow is fixed at $d=64$.

Each baseline model is trained for 1000 epochs, annealing the KL term in the objective function over the first 250 epochs as in \citet{bowman_generating_2016,sonderby_ladder_2016}. The gradient boosted models apply the same training schedule to each component. We optimize using the Adam optimizer \citep{kingma_adam_2015} with a learning rate of $1e-3$ (decay of 0.5x with a patience of 250 steps). To evaluate the negative log-likelihood (NLL) we use importance sampling (as proposed in \citet{rezende_stochastic_2014}) with 2000 importance samples. To ensure a fair comparison, the reported ELBO for GBNF models is computed by \eqref{eq:nf_elbo}---effectively dropping GBNF's fixed components term and setting the entropy regularization to $\lambda=1.0$.

\paragraph{Model Architectures} In Section \ref{sec:real_results}, we compute results on real datasets for the VAE and VAEs with a flow-based approximate posterior. In each model we use convolutional layers, where convolutional layers follow the PyTorch convention \citep{paszke_automatic_2017}. The encoder of these networks contains the following layers:
\begin{align*}
& \mathrm{Conv(in=1, out=16, k=5, p=2, s=2)} \notag \\
& \mathrm{Conv(in=16, out=32, k=5, p=2, s=2)} \notag \\
& \mathrm{Conv(in=32, out=256, k=7, p=0, s=1)} \notag 
\end{align*}
where $\mathrm{k}$ is a kernel size, $\mathrm{p}$ is a padding size, and $\mathrm{s}$ is a stride size. The final convolutional layer is followed by a fully-connected layer that outputs parameters for the diagonal Gaussian distribution and amortized parameters of the flows (depending on model).

Similarly, the decoder mirrors the encoder using the following transposed convolutions:
\begin{align*}
& \mathrm{ConvT(in=64, out=32, k=7, p=0, s=2)} \notag \\
& \mathrm{ConvT(in=32, out=16, k=5, p=0, s=2)} \notag \\
& \mathrm{ConvT(in=16, out=16, k=5, p=1, s=1, op=1)} \notag \\
\end{align*}
where $\mathrm{op}$ is an outer padding. The decoders final layer is passed to standard 2-dimensional convolutional layer to reconstruction the output, whereas the other convolutional layers listed above implement a gated action function:
\begin{equation*}
\mathbf{h}_{l} = ( \mathbf{W}_{l} * \mathbf{h}_{l-1} + \mathbf{b}_{l} ) \odot \sigma ( \mathbf{V}_{l} * \mathbf{h}_{l-1} + \mathbf{c}_{l} ) ,
\end{equation*}
where $\mathbf{h}_{l-1}$ and $\mathbf{h}_{l}$ are inputs and outputs of the $l$-th layer, respectively, $\mathbf{W}_{l}, \mathbf{V}_{l}$ are weights of the $l$-th layer, $\mathbf{b}_{l}, \mathbf{c}_{l}$ denote biases, $*$ is the convolution operator, $\sigma(\cdot)$ is the sigmoid activation function, and $\odot$ is an element-wise product.

\subsection{Density Estimation on Real Data}

\paragraph{Dataset} For the unconditional density estimation experiments we follow \citet{papamakarios_masked_2017,uria_rnade_2013}, evaluating on four dataset from the UCI machine learning repository \citep{dua_uci_2017} and patches of natural images from natural images \citep{martin_database_2001}. From the UCI repository the POWER dataset ($d=6$, $N=$2,049,280) contains electric power consumption in a household over a period of four years, GAS ($d=8$, $N=$1,052,065) contains logs of chemical sensors exposed to a mixture of gases, HEPMASS ($d=21$, $N=$525,123) contains Monte Carlo simulations from high energy physics experiments, MINIBOONE ($d=43$, $N=$36,488) contains electron neutrino and muon neutrino examples. Lastly we evaluate on BSDS300, a dataset ($d=63$, $N=$1,300,000) of patches of images from the homonym dataset. Each dataset is preprocessed following \citet{papamakarios_masked_2017}.

\paragraph{Experimental Setup} We compare our results against Glow \citep{kingma_glow_2018}, and RealNVP \citep{dinh_density_2017}. We train models using a small grid search on the depth of the flows $K \in \{5, 10 \}$, the number of hidden units in the coupling layers $H \in \{ 10d, 20d, 40d \}$, where $d$ is the input dimension of the data-points. We trained using a cosine learning rate schedule with the learning rate determined using the learning rate range test \cite{smith_cyclical_2017} for each dataset, and similar to \citet{durkan_neural_2019} we use batch sizes of 512 and up to 400,000 training steps, stopping training early after 50 epochs without improvement. The log-likelihood calculation for GBNF follows \eqref{eq:gbnf_de_model}, that is we recursively compute and combine log-likelihoods for each component.

\section{Additional Results}
\subsection{Density Estimation}

%
%
\begin{figure}[h]
\centering

\begin{subfigure}[b]{12mm}
\centering
\caption*{\underline{C $\times$ K}}
\vspace{3em}
8 $\times$ 1
\vspace{2em}
\end{subfigure}%
\begin{subfigure}[b]{\figwidth}
\centering
\caption*{Data}
\vspace{-2mm}
\includegraphics[width=\linewidth]{images/density_sampling/8gaussians_samples.jpg}
\end{subfigure}%
\hspace{1mm}%
\begin{subfigure}[b]{\figwidth}
\centering
\caption*{\centering RealNVP (K=8)}
\vspace{-2mm}
\includegraphics[width=\linewidth]{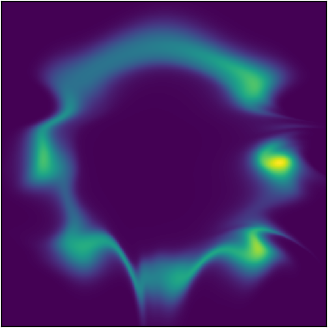}
\end{subfigure}%
\hspace{1mm}%
\begin{subfigure}[b]{\figwidth}
\centering
\caption*{\centering Single Component}
\vspace{-2mm}
\includegraphics[width=\linewidth]{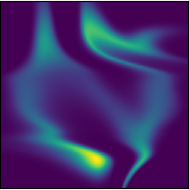}
\end{subfigure}%
\hspace{1mm}%
\begin{subfigure}[b]{\figwidth}
\centering
\caption*{\centering \textbf{Gradient Boosted}}
\vspace{-2mm}
\includegraphics[width=\linewidth]{images/density_sampling/8gaussians_boosted_K1_bs64_C8_reg80_realnvp_tanh1_hsize256.png}
\end{subfigure}%

\vspace{1em}
\begin{subfigure}[b]{12mm}
\centering
4 $\times$ 2
\vspace{2em}
\end{subfigure}%
\begin{subfigure}[b]{\figwidth}
\centering
\vspace{-2mm}
\includegraphics[width=\linewidth]{images/density_sampling/pinwheel_samples.png}
\end{subfigure}%
\hspace{1mm}%
\begin{subfigure}[b]{\figwidth}
\centering
\vspace{-2mm}
\includegraphics[width=\linewidth]{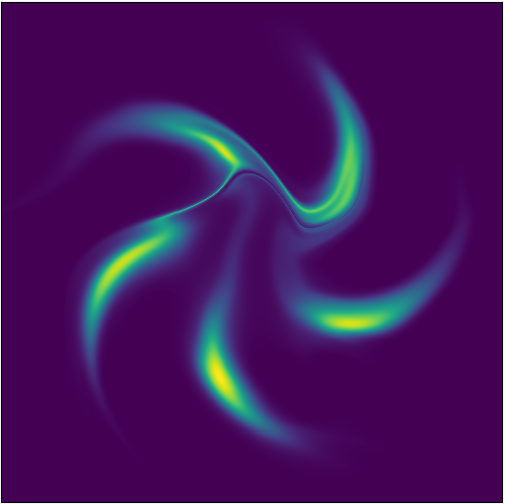}
\end{subfigure}%
\hspace{1mm}%
\begin{subfigure}[b]{\figwidth}
\centering
\vspace{-2mm}
\includegraphics[width=\linewidth]{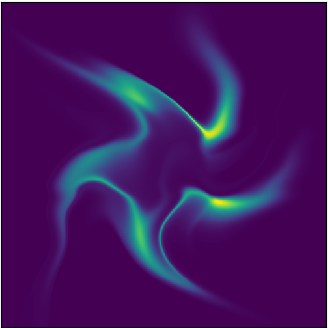}
\end{subfigure}%
\hspace{1mm}%
\begin{subfigure}[b]{\figwidth}
\centering
\vspace{-2mm}
\includegraphics[width=\linewidth]{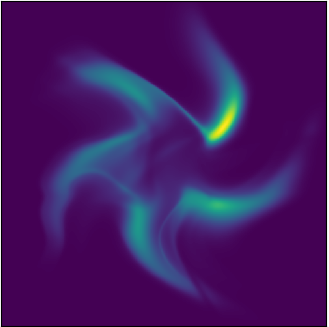}
\end{subfigure}%

\vspace{1em}
\begin{subfigure}[b]{12mm}
\centering
2 $\times$ 4
\vspace{2em}
\end{subfigure}%
\begin{subfigure}[b]{\figwidth}
\centering
\vspace{-2mm}
\includegraphics[width=\linewidth]{images/density_sampling/checkerboard_samples.jpg}
\end{subfigure}%
\hspace{1mm}%
\begin{subfigure}[b]{\figwidth}
\centering
\vspace{-2mm}
\includegraphics[width=\linewidth]{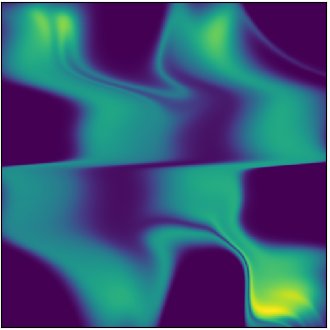}
\end{subfigure}%
\hspace{1mm}%
\begin{subfigure}[b]{\figwidth}
\centering
\vspace{-2mm}
\includegraphics[width=\linewidth]{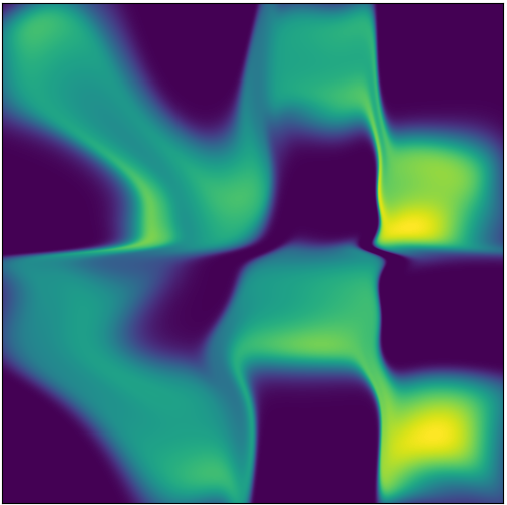}
\end{subfigure}%
\hspace{1mm}%
\begin{subfigure}[b]{\figwidth}
\centering
\vspace{-2mm}
\includegraphics[width=\linewidth]{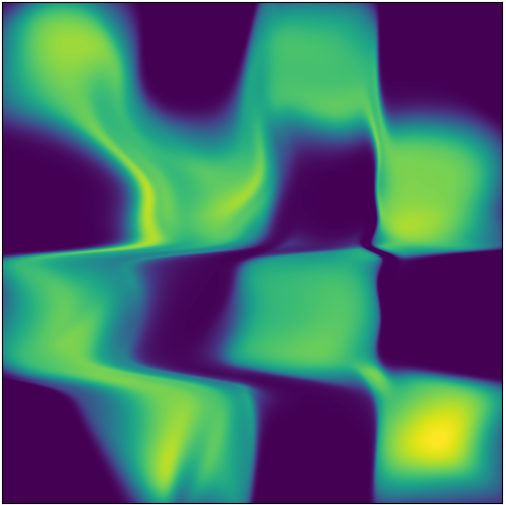}
\end{subfigure}%
\caption{Density estimation for 2D toy data with various settings of components $C$ and flow length $K$. The second column shows a RealNVP model with $K=8$ flows \cite{dinh_density_2017}. In the final column is an equivalently sized Gradient Boosted Flow, consisting of $C$ components each with flow length $K$ (listed to the left of each row). For reference, GBNF's first component (trained using standard methods) is shown in column three --- highlighting how flows of length $K= 1, 2,$ or $4$ can be combined for a more refined and flexible model.}
\label{fig:density_sampling2}
\end{figure}

\subsection{Image Reconstructions from VAE Models}
\newcommand{\reconfigwidth}{0.27\linewidth}
\begin{figure}[t]
\centering

\begin{subfigure}[b]{18mm}
\centering
\caption*{\underline{Dataset}}
\vspace{5em}
Freyfaces
\vspace{6em}
\end{subfigure}%
\begin{subfigure}[b]{\reconfigwidth}
\centering
\caption*{Real}
\includegraphics[width=\linewidth]{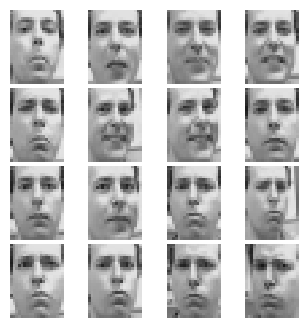}
\end{subfigure}%
\hspace{1mm}%
\begin{subfigure}[b]{\reconfigwidth}
\centering
\caption*{\centering Sylvester}
\includegraphics[width=\linewidth]{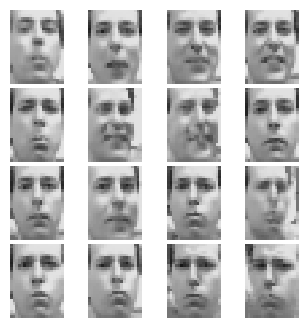}
\end{subfigure}%
\hspace{1mm}%
\begin{subfigure}[b]{\reconfigwidth}
\centering
\caption*{\centering \textbf{Gradient Boosted}}
\includegraphics[width=\linewidth]{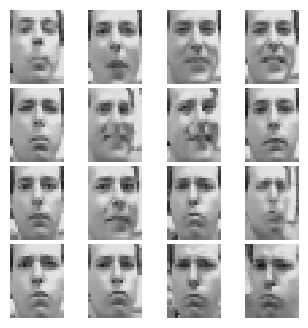}
\end{subfigure}%

\vspace{1em}
\begin{subfigure}[b]{18mm}
\centering
Caltech
\vspace{4em}
\end{subfigure}%
\begin{subfigure}[b]{\reconfigwidth}
\centering
\vspace{-2mm}
\includegraphics[width=\linewidth]{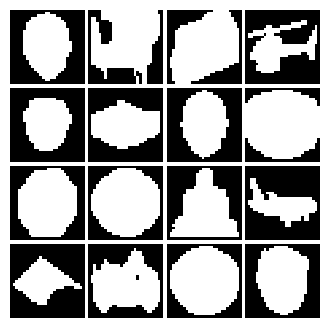}
\end{subfigure}%
\hspace{1mm}%
\begin{subfigure}[b]{\reconfigwidth}
\centering
\vspace{-2mm}
\includegraphics[width=\linewidth]{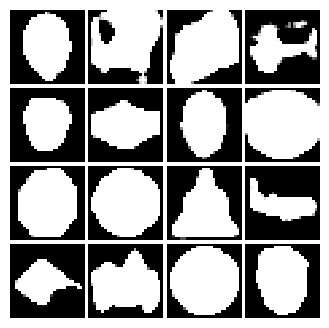}
\end{subfigure}%
\hspace{1mm}%
\begin{subfigure}[b]{\reconfigwidth}
\centering
\vspace{-2mm}
\includegraphics[width=\linewidth]{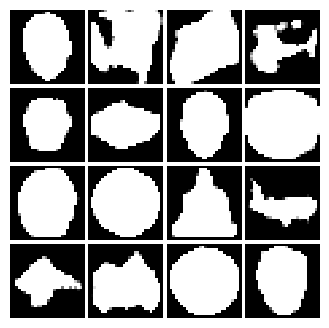}
\end{subfigure}%

\vspace{1em}
\begin{subfigure}[b]{18mm}
\centering
Omniglot
\vspace{4em}
\end{subfigure}%
\begin{subfigure}[b]{\reconfigwidth}
\centering
\vspace{-2mm}
\includegraphics[width=\linewidth]{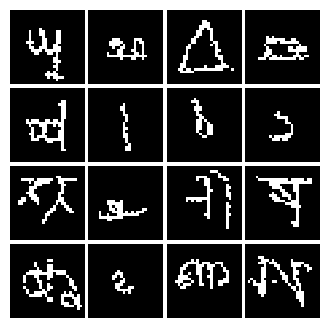}
\end{subfigure}%
\hspace{1mm}%
\begin{subfigure}[b]{\reconfigwidth}
\centering
\vspace{-2mm}
\includegraphics[width=\linewidth]{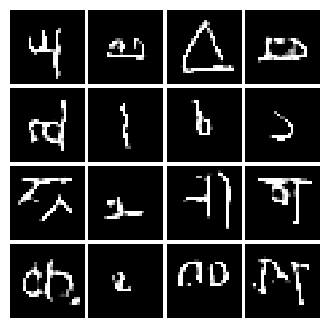}
\end{subfigure}%
\hspace{1mm}%
\begin{subfigure}[b]{\reconfigwidth}
\centering
\vspace{-2mm}
\includegraphics[width=\linewidth]{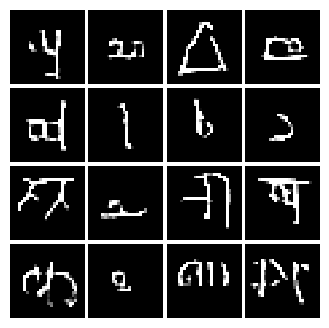}
\end{subfigure}%

\vspace{1em}
\begin{subfigure}[b]{18mm}
\centering
MNIST
\vspace{4em}
\end{subfigure}%
\begin{subfigure}[b]{\reconfigwidth}
\centering
\vspace{-2mm}
\includegraphics[width=\linewidth]{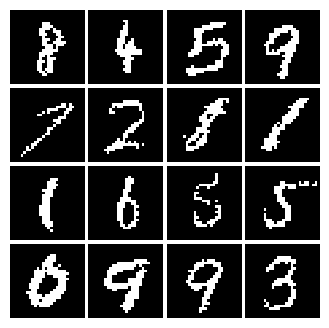}
\end{subfigure}%
\hspace{1mm}%
\begin{subfigure}[b]{\reconfigwidth}
\centering
\vspace{-2mm}
\includegraphics[width=\linewidth]{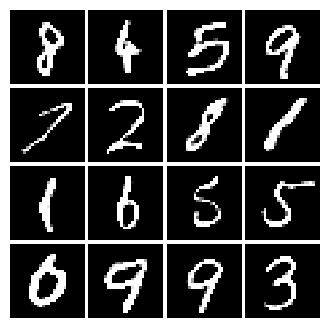}
\end{subfigure}%
\hspace{1mm}%
\begin{subfigure}[b]{\reconfigwidth}
\centering
\vspace{-2mm}
\includegraphics[width=\linewidth]{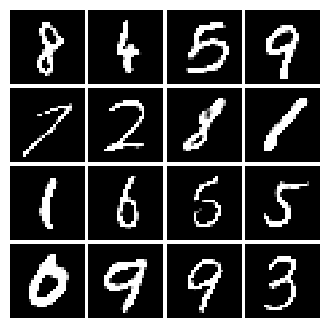}
\end{subfigure}%
\caption[Image Reconstruction for GBF Augments VAE]{Image reconstructions for the four datasets listed in Table \ref{tab:vae}.}
\label{fig:vae_recon}
\end{figure}

\end{document}